\documentclass[journal=jacsat,manuscript=article,layout=twocolumn]{achemso}
\usepackage[version=3]{mhchem} 
\usepackage{amssymb}
\usepackage{booktabs}
\usepackage{tabularx}
\usepackage{bm}

\author{Yulia Pimonova}
\affiliation{Computing and Artificial Intelligence Division, Los Alamos National Laboratory, Los Alamos, NM 87545, USA}
\email{ypimonova@lanl.gov}
\author{Michael G. Taylor}
\affiliation{Theoretical Division, Los Alamos National Laboratory, Los Alamos, NM 87545, USA}
\author{Alice Allen}
\affiliation{Theoretical Division, Los Alamos National Laboratory, Los Alamos, NM 87545, USA}
\alsoaffiliation{Center for Nonlinear Studies, Los Alamos National Laboratory, Los Alamos, NM 87545, USA}
\altaffiliation{Max Planck Institute for Polymer Research, Ackermannweg 10, 55128 Mainz, Germany}
\author{Ping Yang}
\email{pyang@lanl.gov}
\affiliation{Theoretical Division, Los Alamos National Laboratory, Los Alamos, NM 87545, USA}
\author{Nicholas Lubbers}
\affiliation{Computing and Artificial Intelligence Division, Los Alamos National Laboratory, Los Alamos, NM 87545, USA}
\email{nlubbers@lanl.gov}

\title[mlearner]
  {Meta-Learning Linear Models for Molecular Property Prediction}

\abbreviations{}
\keywords{}

\begin{document}

\begin{tocentry}
\centering
\includegraphics[width=8.3cm]{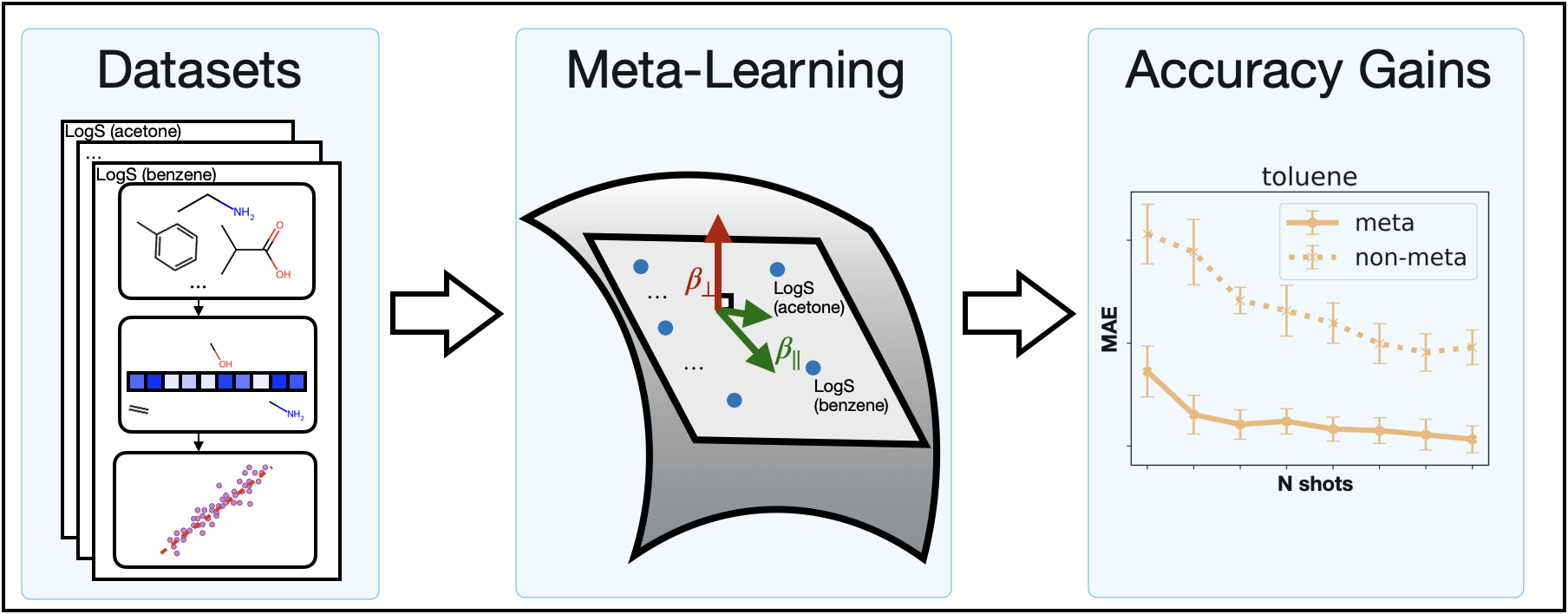}
\label{fig:toc}
\end{tocentry}

\begin{abstract}

Chemists in search of structure-property relationships face great challenges due to limited high quality, concordant datasets. Machine learning (ML) has significantly advanced predictive capabilities in chemical sciences, but these modern data-driven approaches have increased the demand for data. In response to the growing demand for explainable AI (XAI) and to bridge the gap between predictive accuracy and human comprehensibility, we introduce \emph{LAMeL}—a Linear Algorithm for Meta-Learning that preserves interpretability while improving the prediction accuracy across multiple properties. While most approaches treat each chemical prediction task in isolation, \emph{LAMeL} leverages a meta-learning framework to identify shared model parameters across related tasks, even if those tasks do not share data, allowing it to learn a common functional manifold that serves as a more informed starting point for new unseen tasks. Our method delivers performance improvements ranging from 1.1- to 25-fold over standard ridge regression, depending on the domain of the dataset. While the degree of performance enhancement varies across tasks, \emph{LAMeL} consistently outperforms or matches traditional linear methods, making it a reliable tool for chemical property prediction where both accuracy and interpretability are critical.

\end{abstract}

\section{Introduction}
Machine learning has transformed how we approach complex chemical problems, delivering accurate predictions for molecular properties and chemical reactivity while reducing the need for both costly experimental evaluations and more affordable computational methods.~\cite{keithCombiningMachineLearning2021, lanNewInsightsPredictions2022, aldossarySilicoChemicalExperiments2024} The intersection of machine learning and chemistry manifests in new transformative approaches to molecular property prediction, materials design, and reaction conditions optimization.~\cite{marimuthu2025, zhengLargeLanguageModels2025, liWhenQuantumMechanical2024, burrillMLTBEnhancingTransferability2025} Over the years, increasingly sophisticated ML algorithms have demonstrated remarkable success in predicting chemical properties, ranging from solubility and drug-likeness to atomization energies and reaction dynamics.~\cite{aminghanavatiMachineLearningApproach2024, kyroChemSpaceALEfficientActive2024, fedikChallengesOpportunitiesMachine2025, zhangIncludingPhysicsInformedAtomization2025} Despite these advances, the chemistry domain, along with other physical sciences, presents unique challenges that many common ML approaches struggle to fully address.

Modern chemical machine learning faces constant competition between predictive power and interpretability. Deep neural networks, graph neural networks, and other complex architectures have achieved state-of-the-art performance across numerous chemical prediction tasks.~\cite{yangPredictingChemicalShifts2021, heidChempropMachineLearning2024, manstineAIMNet2NeuralNetwork2025} Nevertheless, these models function largely as ``black boxes,''  making their decision-making processes opaque to human understanding.~\cite{interpretcall_rudin2019} This interpretability challenge is particularly acute in chemistry, where understanding the underlying structure-property relationships is essential to continuous scientific innovation. Chemists have traditionally relied on transparent and mechanistically meaningful models that reveal how specific structural features influence molecular properties.~\cite{wellawatteHumanInterpretableStructureproperty2025}

On the other hand, linear models are inherently interpretable, which stems from their explicit parameter weights.  The coefficients in linear models directly quantify the contribution of each feature, allowing for direct interpretation of prediction results.
Although linear models often lag behind neural networks in terms of performance, their transparency and ease of interpretation are compelling incentives to use them, even when they are less accurate.~\cite{venturaComparisonMultipleLinear2013antitub, caiQuantitativeStructureActivity2022} Recent contribution by Allen and Tkatchenko demonstrates that, with an appropriate featurization scheme, multi-linear regression can achieve performance comparable to more advanced deep learning architectures in predicting materials properties.~\cite{multilinearAllen2022} Moreover, linear regression models are faster than neural networks in terms of both training speed and computational resource requirements due to the their much simpler design.

The widespread application of ML in the physical sciences meets a major challenge in the pervasive data scarcity in experimental studies. Acquiring chemical data—and, more broadly, any experimental data—is resource-intensive, time-consuming, and expensive. The problem is especially pronounced in drug design~\cite{altae-tranLowDataDrug2017, vellaFewShotLearningLowData2023a,zhangImprovedPredictionDrug2025} but extends across many areas of chemistry~\cite{LowDataChemicalDatasets_Tom, haasApplyingStatisticalModeling2025}. When experimental data are scarce, combining low-fidelity simulation with limited high-fidelity experiments can improve accuracy and robustness, as shown by Nevolianis et al. for toluene–water partition coefficients~\cite{nevolianisMultifidelityGraphNeural2025}. Similarly, Eraqi et al. have demonstrated that multi-task learning over multiple sustainable aviation fuel properties provided benefits in the ultra-low data regime with as few as 29 samples.~\cite{eraqiMolecularPropertyPrediction2025} The low-data problem becomes particularly critical when the demand for high-accuracy prediction is high.~\cite{zhangImprovedPredictionDrug2025, singhMetalearningApproachSelectivity2025}

Meta-learning has emerged as a powerful framework to address data efficiency challenges across diverse machine learning domains. Unlike methods that treat each task independently, meta-learning seeks to ``learn to learn'' by leveraging shared structure across related tasks~\cite{vettoruzzoAdvancesChallengesMetaLearning2023, huismanSurveyDeepMetalearning2021}. This paradigm enables models to acquire transferable knowledge that facilitates rapid adaptation to new tasks, even in low-data regimes. Meta-learning distinguishes itself from other knowledge transfer frameworks such as transfer learning and multitask learning.\cite{vettoruzzoAdvancesChallengesMetaLearning2023} While multitask learning focuses on simultaneously learning multiple tasks to perform well on those same tasks,~\cite{caruanaMultitaskLearning1997} meta-learning is designed to ``learn how to learn,'' enabling models to quickly adapt to entirely new tasks with minimal examples. This contrasts with transfer learning, which leverages knowledge from previously learned source tasks to enhance performance on a different target task through fine-tuning.~\cite{zhuangComprehensiveSurveyTransfer2021, smithApproachingCoupledCluster2019} The key distinction of meta-learning lies in emphasis on rapid adaptation to new tasks rather than just applying existing knowledge (transfer learning) or handling multiple known tasks concurrently (multitask learning). Meta-learning develops a learning capability that allows models to efficiently learn new information with few training examples. Recent studies have demonstrated the promise of meta-learning in chemistry-related applications. For instance, Allen et al.~\cite{allenLearningTogetherFoundation2024} showed that incorporating multiple levels of quantum chemical theory within a unified training process can enhance prediction accuracy. Building on this promise, Wang et al.~\cite{wangFoundationModelChemical2024} integrated meta-learning into the design of a foundation model for chemical reactors, while Singh and Hernández-Lobato applied prototypical networks~\cite{NIPS2017_cb8da676} to improve selectivity predictions along organic reaction pathways.~\cite{singhMetalearningApproachSelectivity2025}

Despite these advances, most existing meta-learning approaches emphasize deep learning architectures that prioritize predictive performance at the expense of interpretability. Qian et al.~\cite{qianMetaLearningAttention2024a} specifically highlight the lack of interpretability as a major limitation in their few-shot molecular property prediction model. In response, several efforts have aimed to improve interpretability. One strategy involves developing interpretable models that replicate the performance of deep networks, such as the approach proposed by Fabra-Boluda et al.~\cite{fabra-boluda_cracking_2025}. More commonly, post-hoc interpretability techniques are employed. These include symbolic metamodels layered on top of neural networks~\cite{alaa_demystifying_2019}, analyses of specific hidden layers~\cite{liu_interpretable_2018}, regression models based on architectural meta-features~\cite{pereira_neural_2023}, and variance decomposition methods such as Meta-ANOVA~\cite{choiMetaanovaScreeningInteractions2025}.

This limitation highlights a critical knowledge gap: the absence of application-oriented meta-learning algorithms specifically designed for linear models. While there is growing academic interest in this area, most existing efforts remain theoretical and lack practical application to real-world problems. For instance, Tripuraneni et al. introduced a provably sample-efficient algorithm for multi-task linear regression, focusing on learning shared low-dimensional representations across tasks.~\cite{tripuraneniProvableMetaLearningLinear2022} While their contribution offers strong theoretical proof, it does not address practical deployment challenges. Similarly, Denevi et al. proposed a conditional meta-learning approach that tailors representations to individual tasks using side information, offering improved adaptation in clustered task environments, yet their method has not been tested in applied settings.~\cite{deneviConditionalMetaLearningLinear2021}  Toso et al. extended meta-learning to linear quadratic regulators using a model-agnostic approach, demonstrating theoretical guarantees for controller stability and adaptation, but their focus remains on control theory without broader application.~\cite{tosoMetaLearningLinearQuadratic2024}  These studies underscore the need for developing meta-learning algorithms for linear models that are not only theoretically sound but also practically applicable across diverse real-world domains.

\begin{figure*}[h!]
    \centering
    \includegraphics[width=0.95\textwidth]{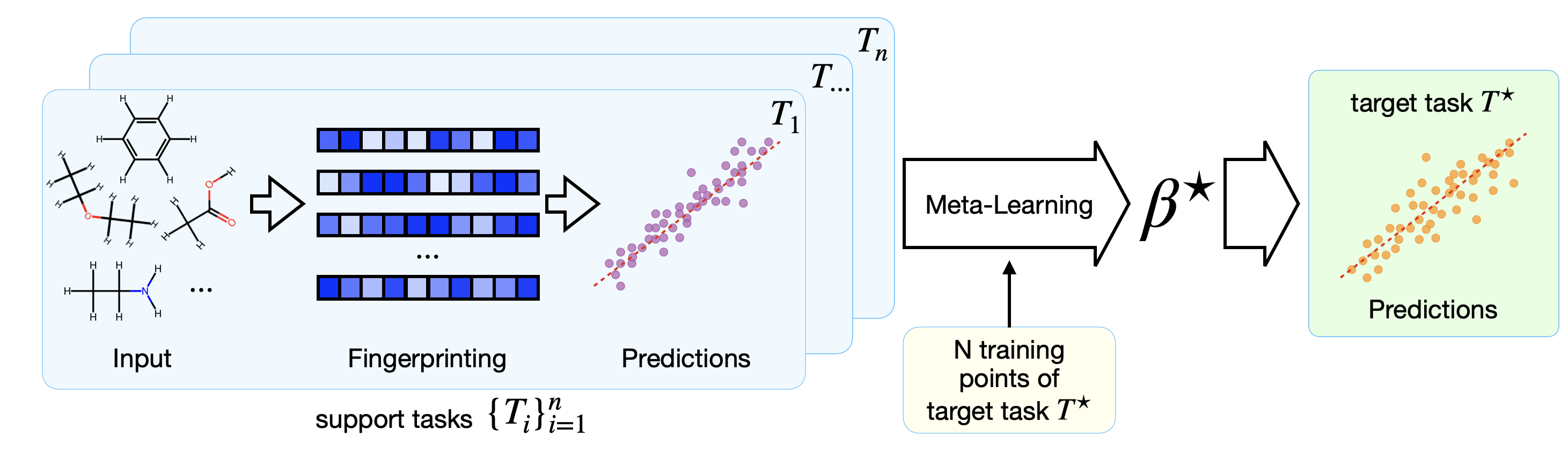}
    \caption{Overview of the LAMeL workflow. Molecular structures are first converted into numerical representations that serve as features for predictive modeling. Support tasks $( \{ T_i \}_{i=1}^n )$ are used in a meta-learning framework to identify shared parameters across related tasks. These shared parameters provide an informed initialization for few-shot learning on a new target task $T^{\star}$, enabling accurate predictions with limited data.}
    \label{fgr:workflow}
\end{figure*}

To bridge the gap in applying meta-learning to linear models for chemical domain, we introduce \emph{LAMeL}—a novel algorithm that reshapes meta-learning principles specifically for linear architectures. \emph{LAMeL} learns shared parameters across related support tasks, identifying a common functional manifold that serves as an informed initialization for new, unseen tasks. The familiarized starting point enables the meta-model to adapt to new tasks with only a few data points. Fig.~\ref{fgr:workflow} illustrates the LAMeL workflow, showcasing how meta-learning is applied to linear models for chemical property prediction by leveraging support tasks to enhance performance on a target task.

The primary contributions of this work include:
\begin{itemize}
    \item The development of LAMeL, the first meta-learning algorithm specifically designed for linear models in chemistry applications.
    \item A comprehensive validation of LAMeL across multiple chemical domains, demonstrating performance improvements ranging from 1.1 to 25-fold over classical ridge regression.
    \item An investigation into the role and importance of the task similarity across support data. 
\end{itemize}
By providing an ML tool that preserves interpretability while working in the low-data regime, LAMeL contributes to the broader goal of making ML-acquired results more valuable for chemistry.

\section{Methods}

\subsection{Dataset Descriptions}
\emph{Boobier Solubility Prediction Dataset}

The dataset developed by Boobier et al.~\cite{boobierMachineLearningPhysicochemical2020} integrates experimental solubility across four solvent systems: water, acetone, benzene and ethanol. The dataset presents a unique opportunity for us to access not only the viability of our meta-learning approach, but also get a sense of task similarity effects on the \emph{LAMeL} prediction accuracy. The reported prediction accuracy range of 60-80\% within LogS±0.7 demonstrates state-of-the-art performance for physics-informed solubility models.

\emph{BigSolDB 2.0 Solubility Prediction Dataset}

BigSolDB 2.0~\cite{krasnovBigSolDB20Dataset2025a} is a large, openly accessible dataset that compiles experimental solubility measurements for organic compounds. BigSolDB 2.0 aggregates 103,944 solubility values from 1,595 studies, creating one of the largest repository for non-aqueous solubility prediction. The dataset spans 1,448 unique organic solutes and 213 solvents, with temperature-dependent measurements the range of 243–425 K. 
Each entry contains structures of solutes and solvents (as SMILES strings), experimental solubility values (as log-values of molarity), temperature, and bibliographic information for the originating study. The breadth and diversity of BigSolDB 2.0 make it a valuable source for benchmarking ML models of solubility. 

\emph{QM9-MultiXC Molecular Energy Dataset}

The QM9-MultiXC~\cite{nandiMultiXCQM9LargeDataset2023} is an extension to the popular QM9 dataset~\cite{QM9_2014}. It provides a systematic comparison of quantum chemical methods through 228 distinct energy calculations per molecule, incorporating 76 density functional theory (DFT) functionals combined with three different basis sets (SZ, DZP, and TZP) of varying fidelity. For linear meta-models, this multi-fidelity approach enables systematic investigation of theoretical method dependencies, particularly through the analysis of prediction power of the meta-learning approach across different combinations of computational approaches as support tasks. The philosophy of meta-learning supports development of transferable and interpretable structure-energy relationships, creating a controlled testbed for functional transfer learning across theory levels.

\subsection{Substructural Fingerprints}
In cheminformatics, molecular representation is a critical step in translating chemical structures into a format suitable for regression tasks. There are numerous ways to accomplish this process, commonly referred to as \emph{fingerprinting} the molecules~\cite{wighReviewMolecularRepresentation2022}.  In this work, we use graphlet fingerprints, a direct topological representation that has demonstrated effectiveness in molecular property prediction.~\cite{Tynes2024} Graphlet fingerprints have emerged as a powerful tool for molecular representation in cheminformatics, offering a balance between structural granularity and computational efficiency.~\cite{Tynes2024, exhuastive_graphlets}.

\begin{figure*}[h!]
    \centering
    \includegraphics[width=0.85\textwidth]{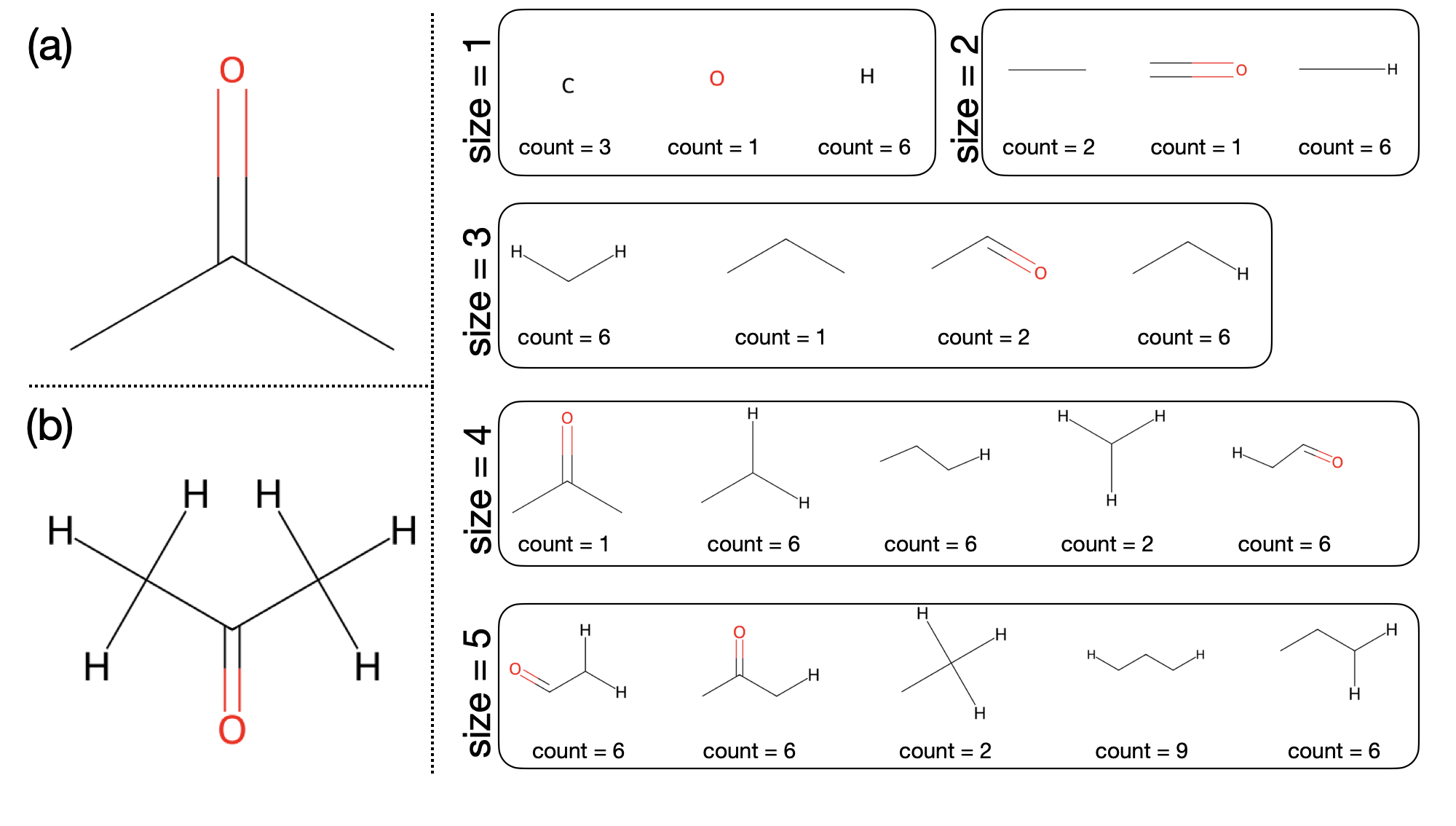}
    \caption{Graphlet decomposition of acetone up to a maximum substructure size of 5. (a) Conventional molecular graph representation of acetone with implicit hydrogens. (b) Full molecular graph of acetone with explicit hydrogens. (c) Enumeration of graphlet substructures, grouped by size. Each box contains the unique graphlets identified at a given size, along with their occurrence count.}
    \label{fgr:minerva_example}
\end{figure*}

Graphlets operate on a the molecular graph in which atoms serve as nodes and bonds as edges. Graphlets are formed from the isomorphism classes of connected subgraphs in the molecular graph; a one-node graphlet constitutes a single atom, a two-node graphlet constitutes two bonded atoms and the associated bond type, and so on for higher-sized molecular fragments. Using graphlet representations in molecular property prediction builds upon the many-body expansion principle in quantum chemistry,~\cite{richardGeneralizedManybodyExpansion2012} where properties are approximated as sums of contributions from increasingly complex atomic clusters. The fingerprinting process involves systematically enumerating all graphlets within a molecule up to a predefined maximum graphlet size. Fig.~\ref{fgr:minerva_example} illustrates this process for acetone, where all graphlets up to size 5 are extracted from the molecular graph. Unlike path-based\cite{rdkit} or radial fingerprints~\cite{morganGenerationUniqueMachine1965}, graphlets capture every kind of substructure, provide a more complete encoding of molecular topology as they identify every possible subgraph.  A fast, recursive hashing procedure allows identifying the isomorphism class of each graphlet. 

The set of all graphlets in a given dataset can then be assembled as a feature matrix, giving counts of each type of substructure in each molecule in the dataset. This preserves an  interpretable relationship between molecular components and their contributions to predicted properties.
In our meta-learning framework, model coefficients correspond directly to specific graphlet substructures, and as a result, meta-learned models preserve the interpretability of the graphlet featurization approach. It stands to reason that the structured organization of the features might facilitate knowledge transfer across tasks as it mimics the structure that human chemists use to build chemical intuition.

\subsection{Meta-Learning Background}

Meta-learning, often described as ``learning to learn,'' is an approach in machine learning where the goal is to develop models that can rapidly adapt to new tasks by leveraging experience gained from a collection of related tasks. In the meta-learning framework, we typically distinguish between \emph{support tasks} $T_i$, $i \in \{1\dots T\}$ - a set of training tasks used to learn meta-knowledge - and \emph{target tasks} $T^\star$, which are novel tasks where the meta-learned knowledge is applied to achieve fast adaptation and improved performance.

Each task $T_i$ is characterized by its own dataset, consisting of task features $X^T$ and corresponding labels $y^T$. The meta-learning is advantageous over other knowledge-transfer approaches as it does not need the same data point to appear across multiple tasks; rather, each task can have its own unique data distribution and labeling. This flexibility enables meta-learning to handle a diverse range of tasks with sparsely distributed data.

The meta-learning algorithm presented in this work can be considered an optimization-based approach, where the meta-learner aims to find model or initialization parameters that facilitate rapid adaptation to new tasks with minimal data. During meta-training, the model is exposed to multiple support tasks, learning to optimize its parameters such that, when presented with a target task, it can quickly fine-tune to achieve the best performance with minimal data. This approach is particularly effective in scenarios characterized by limited labeled data for new tasks. In such cases, the target task is inherently data-constrained, with only a small number of data points available for training. We refer to these data points as \emph{shots}, following the convention established in the few-shot learning literature,~\cite{songComprehensiveSurveyFewshot2023, gharounMetalearningApproachesFewShot2024} where the objective is to achieve robust generalization from a minimal number of training examples.

\subsection{\emph{LAMeL}: Linear Algorithm for Meta-Learning }

Our meta-learning approach is aimed at building a linear specialization coefficients $\boldsymbol{\beta}^*$ with predictions $y^*_i = \mathbf{x}_i \cdot \boldsymbol{\beta}^*$.
The model is decomposed into parallel and perpendicular components, that is, $\mathbf{\boldsymbol{\beta}}^*$ = $\boldsymbol{\beta^\perp} + \boldsymbol{\beta^\parallel}$. 
The component $\boldsymbol{\beta^\parallel}$ is in-plane with the $T$-dimensional subspace $W^\parallel$ generated by the support task coefficients $\boldsymbol{\beta}^\tau , \tau \in \{T_1\dots T_T\}$, whereas $\boldsymbol{\beta^\perp}$ is perpendicular to this subspace.
We assume that, as tasks may have some relationship to each other, they may be approximated by a lower-rank manifold.~\cite{tripuraneniProvableMetaLearningLinear2022, ManifoldMetalearningReducedcomplexity} As such, we bias the specialization coefficients $\boldsymbol{\beta}^*$ towards the manifold defined by the models built on support tasks and allow for knowledge distillation from previous learning experiences.

While many forms of bias are possible, we use sequential fitting, which has the advantage of separating out hyperparameter searches. First, we fit within the subspace $W^\parallel$, and subsequently use this start space to find a residual (intuitively, smaller) component $\boldsymbol{\beta^\perp}$. We use a ridge loss function as a base regressor. 

The first step is to build individual support models by minimizing the support task loss
\begin{equation}
\mathcal{L}^\tau = \sum_i (\boldsymbol{x}^\tau_i \cdot \boldsymbol{\beta}^\tau - y^\tau_i)^2 + \lambda^\tau ||\boldsymbol{\beta}^\tau||^2_2. 
\end{equation}
Here, \begin{itemize}
    \item $\|\boldsymbol{\beta}\|_2^2 = \sum_{j=1}^p \beta_j^2$ is the squared L\(_2\) norm of the regression parameters vector,
    \item $\lambda \geq 0$ is the regularization parameter controlling the strength of the penalty,
    \item $\tau $ is each of the support tasks in $\{T_{1},\dots, T_{T}\}$.
\end{itemize}
Ridge regression encourages smaller coefficient values for model's stability and generalization.~\cite{ESL2009}

Next, building features which explore $W^\parallel$ is a matter of dotting the specialization feature matrix $X^*$ with the model matrices, yielding new meta-features, $X^* \cdot \boldsymbol{\beta}^\tau$, which are the prediction of the support tasks applied to the data in the target task. Thus we see explicitly how meta-learning can operate on disjoint support and target tasks: the models from a support task can still be applied to the target task, and we can use the results of these models as features for forming a new prediction.  We build the meta-features using the average support task as the origin for fitting $\boldsymbol{\beta^\parallel}$.
Setting an origin for fitting encourages the learned coefficients $\boldsymbol{\beta^\parallel}$ to stay close to the origin (prior) vector, effectively embedding information from previous learning experiences into the model. This adjustment aligns with the principles of meta-learning, where knowledge from support tasks informs the learning process for a new task. By incorporating a task-specific or meta-learned prior $\boldsymbol{\beta}_{\mathrm{prior}}$, this approach enforces the adaptability of linear models in scenarios with limited available data, as the prior knowledge can mitigate over-fitting and improve generalization and transferability to new tasks.

\begin{figure*}[h!]
    \centering
    \includegraphics[width=0.85\textwidth]{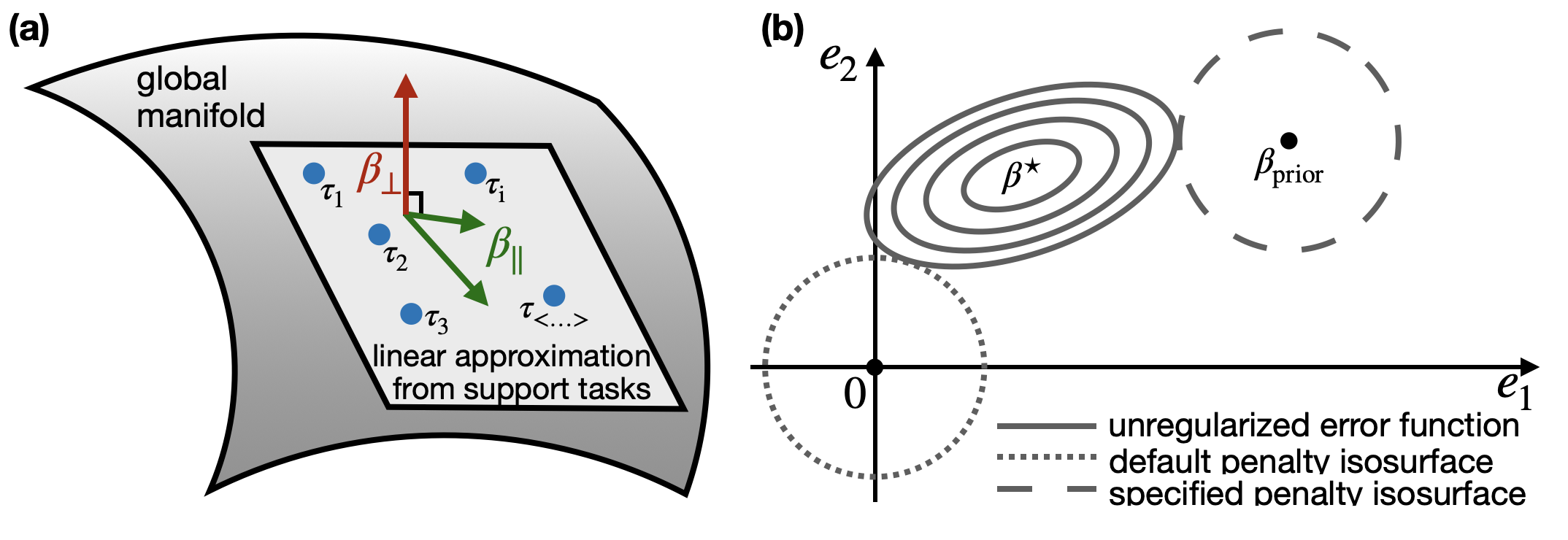}
    \caption{(a) \textit{LAMeL} meta-learning framework for linear models. Support task parameters ($\tau_i \in \textit{T}$) span a local subspace, where the parameter vector for the new task is reconstructed from the parallel (\(\boldsymbol{\beta}_{\parallel}\)) and orthogonal (\(\boldsymbol{\beta}_{\perp}\)) components. (b) The modified L\(_2\) regularization with shifted regularization center.}
    \label{fgr:fit_with_prior}
\end{figure*}

We center these features $\chi^{\tau}_i = (\boldsymbol{\beta}^\tau - \bar{\boldsymbol{\beta}}) \cdot x_i$ using the average support model $\bar{\boldsymbol{\beta}} = \frac{1}{T} \sum_i \boldsymbol{\beta}^t$ as an origin for residual fitting. We then build $\boldsymbol{\beta^\parallel}$ by finding $\mathbf{c} \in \Bbb{R}^T$ minimizing the ridge loss function
\begin{equation}
\mathcal{L}^\parallel = \sum_i (\boldsymbol{\chi}_i \cdot \boldsymbol{c} - (y^*_i - \bar{\hat{y}}_i)^2) + \lambda^\parallel ||\boldsymbol{c}||^2_2,
\end{equation}
where the average-across-tasks prediction is $\bar{\hat{y}}_i = \frac{1}{T} \sum_\tau \hat{y}_i$. This yields the parallel component of the model,
\begin{equation}
    \boldsymbol{\beta}^\parallel = \bar{\boldsymbol{\beta}} + \sum_\tau c^\tau (\boldsymbol{\beta}^\tau - \bar{\boldsymbol{\beta}}). 
\end{equation}
(Note-the loss function $\mathcal{L}^\parallel$ is formally degenerate as there are $T$ coefficients and $T-1$ independent features - however, the resulting $\boldsymbol{\beta}^\parallel$ is well-defined.)
Finally, the residual coefficient $\boldsymbol{\beta}^\perp$ is found by minimizing the ridge loss function of the residuals $\epsilon_i = y_i - x_i \boldsymbol{\beta}^\parallel$ given by
\begin{equation}
\mathcal{L}^\perp = \sum_i (\boldsymbol{x}_i \cdot \boldsymbol{\beta}^\perp - \epsilon_i)^2 + \lambda^\perp ||\boldsymbol{\beta}^\perp||^2_2,
\end{equation}
Summarizing, the three phases of the \emph{LAMeL} algorithm are:
\begin{enumerate}
\item Determine support coefficients $\boldsymbol{\beta}^\tau$ using the support task data, and construct meta-features $\chi_i$ by applying these models to the target task features. 
\item Determine parallel coefficients $\boldsymbol{\beta}^\parallel$ using the meta-features $\boldsymbol{\chi}_i$.
\item Determine perpendicular coefficients $\boldsymbol{\beta}^\perp$ using the ordinary features $\boldsymbol{x}_i$.
\end{enumerate}
The final parameter vector for the specialization task after the few shot learning is:
\begin{equation}
    \mathbf{\boldsymbol{\beta}}^* = \boldsymbol{\beta^\perp} + \boldsymbol{\beta^\parallel}
\end{equation}

\section{Results and Discussion}
We evaluate the proposed meta-learning method on datasets selected to probe generalization and scalability. Our objectives are to test sample efficiency on small problems, assess robustness when the number of prediction tasks is large but per-task supervision is sparse, and analyze performance when both task count and per-task data volume are high. We consider three regimes: a small benchmark with only a few tasks~\cite{boobierMachineLearningPhysicochemical2020}; a large multitask benchmark in which each task provides a single label per sample~\cite{nandiMultiXCQM9LargeDataset2023}; and a large benchmark with many tasks but limited observations per task~\cite{krasnovBigSolDB20Dataset2025a}. This spectrum allows us to examine how the model behaves as the task set grows, as per-task data increases or decreases, and as distributional heterogeneity increases, providing a balanced basis for the comparisons reported below.

\begin{table*}[h]
    \centering
    \renewcommand{\arraystretch}{1.2} 
    \begin{tabular}{cccc} 
        \toprule
         max size & \textit{Boobier et. al.} & \textit{
         BigSol DB 2.0} & \textit{QM9-MultiXC} \\
        \midrule
        3 & 319 & 380 & 125 \\
        5 & 4992 & 5194 & 3280 \\
        7 & 57346 & 58365 & 82942 \\
        \bottomrule
    \end{tabular}
    \caption{Parameter count (number of graphlets) in the topological fingerprints as a function of maximum graphlet size across different datasets.}
    \label{tab:parameter_sizes}
\end{table*}

\subsection{Solubility Database by Boobier et al.}
As described above, the Boobier et al.~\cite{boobierMachineLearningPhysicochemical2020} dataset includes solubility data for the four solvents: water, ethanol, acetone, ethanol. Given only four tasks, in our meta-learning procedure we tested each of the solvents as a target task with the remaining three being used as support tasks. An extra test was performed to ensure the target task data does not leak into the support tasks data used in the meta-learning phase. Detailed results for individual solvents at different maximum substructure sizes used for fingerprinting are provided in Supporting Information. These analyses further illustrate how molecular representation influences prediction accuracy across tasks.

\begin{figure}[h!]
    \centering
    \includegraphics[width=\columnwidth]{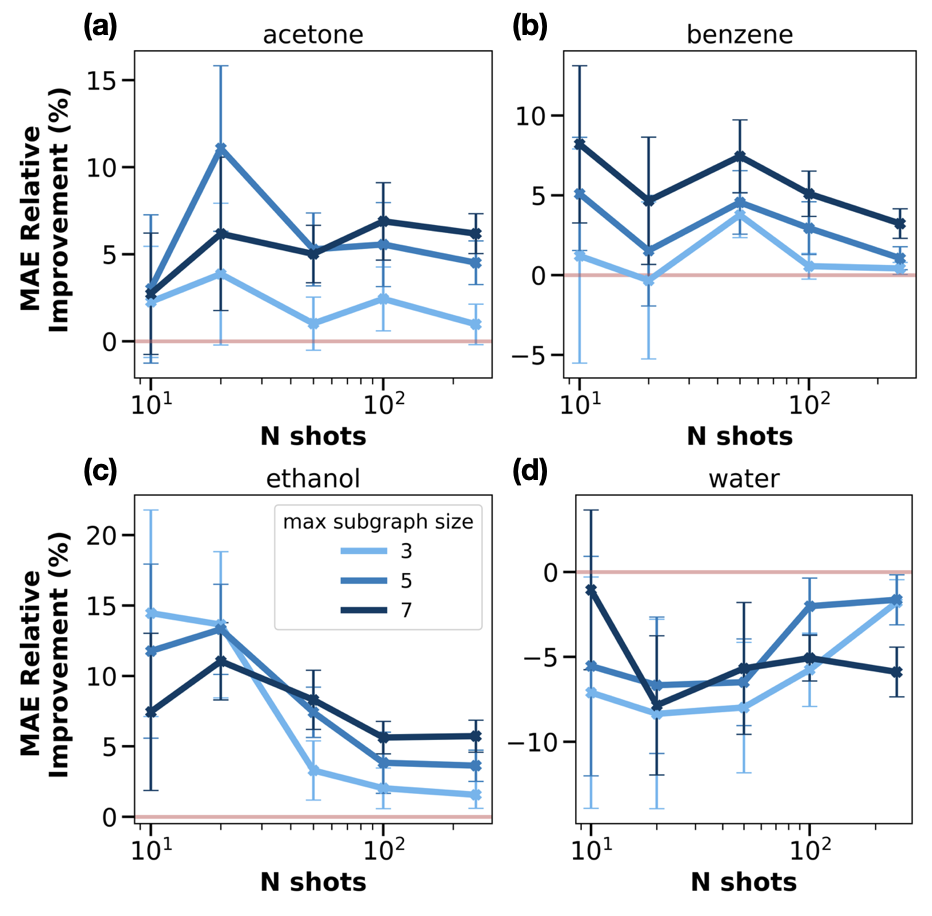}
    \caption{Relative improvement (\%) of meta-learning over non-meta-learning approaches as a function of the number of shots (N shots) and maximum subgraph size for four solvents: (a) acetone, (b) benzene, (c) ethanol, and (d) water. Curves correspond to different maximum subgraph sizes, with error bars indicating standard error over 10 random initializations. Positive values indicate improved performance of \emph{LAMeL} approach.}
    \label{fgr:boobier_all_relative}
\end{figure}

The results of our experiments reveal moderate improvements in prediction accuracy when employing meta-learning, with gains diminishing as the number of shots increases across all target tasks (Fig.~\ref{fgr:boobier_all_relative}). Notably, predictions for water solubility showed no improvement across all shot sizes. We primarily attribute this lack of improvement to the chemical distinctiveness of water compared to the other solvents (ethanol, acetone, and benzene), which serve as support tasks in this case. Water has the highest dielectric constant of ($\varepsilon$=80.1) in comparison with the rest: ethanol ($\varepsilon$=24.5) ,  acetone ($\varepsilon$=20.7)  and benzene is around ($\varepsilon$=2.27).  Water's more polarizing nature and formation of hydrogen-bonding likely reduce its similarity to the other solvents, limiting meta-learning's ability to transfer knowledge from support to target tasks. This observation underscores a key limitation of meta-learning: task similarity among support tasks play an important role in effective knowledge transfer. When support tasks are chemically or structurally dissimilar to the target task, meta-learning exhibits limited ability to exploit common patterns across tasks.

\subsection{BigSolDB 2.0}
Encouraged by the positive results obtained with the Boobier dataset, we applied  meta-learning on the largest solubility dataset, BigSolDB 2.0, comprising approximately ~104K datapoints for 1448 compounds across 213 solvents with temperature variation. In this work we did not want to consider temperature a variable, so in the pre-processing stage for each solvent-solute pair we kept a single entry from the temperature window between 290K and 300K, all other datapoints were disregarded. There are 70 unique solvents in the remaining data. Furthermore, we can filter the solvents (tasks) based on the total number of datapoints per task. To see how the total number of available solvent changes with imposed limitations on the data size, navigate to Fig. S5 in the SI. 

\begin{figure*}[h!]
    \centering
    \includegraphics[width=0.95\textwidth]{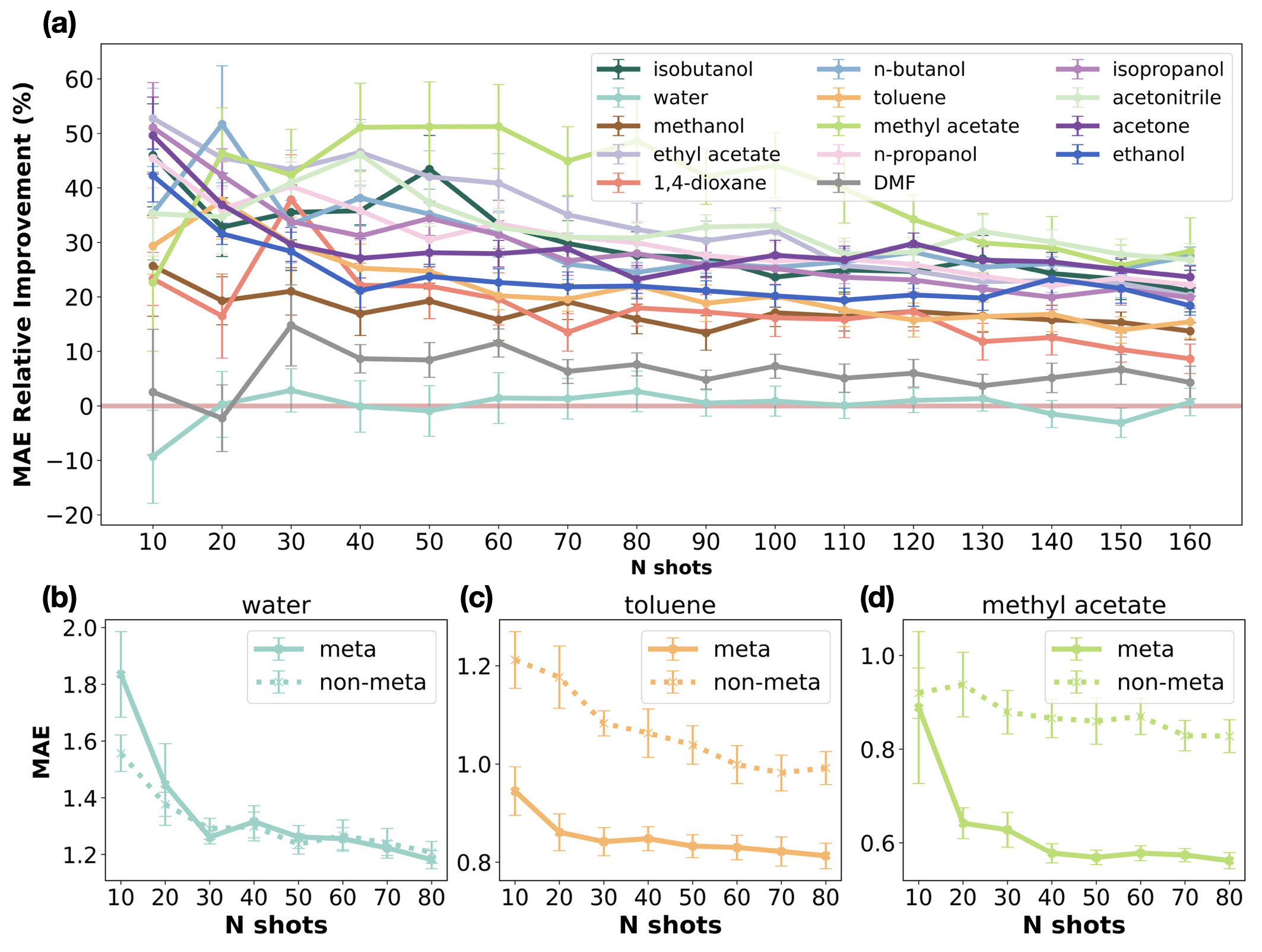}
    \caption{Relative improvement in solubility prediction error (MAE) using meta-learning for BigSolDB 2.0 solvents with a minimum of 200 data points per solvent (a). Panels (b–d) display meta-learning performance for individual solvents—water, toluene, and methyl acetate—ranked by increasing relative improvement.}
    \label{fgr:bsdb_res}
\end{figure*}

As before, in experimental setup all of the tasks that are not the target task have been used as support in the meta-training stage. We observed performance improvement with meta-learning for all but one solvent, water. To quantify the effect we have calculated the relative improvement 
\begin{equation}
    \rm{Rel.\ Imp.}=\frac{\rm{MAE}_{\rm regular} - \rm{MAE}_{\rm meta}}{\rm{MAE}_{\rm meta}} \cdot 100\%
\end{equation} demonstrated in Fig.~\ref{fgr:bsdb_res}. 

In the true few-shots regime (10-30 training datapoints for the target task) solvents exhibit substantial relative improvements, with achieving up to 60\% MAE reduction. This high-variance, high-reward region demonstrates the typical few-shot learning behavior where limited data can yield significant performance gains for well-suited systems. A convergence pattern emerges as relative improvements gradually decrease and stabilize with the increasing number of training points. The high variability observed in the low-shot regime diminishes with more consistent model performance across different solvents. Most solvents converge to a plateau region with relative improvements within 15-30\%.
Similarly to Boobier et al. dataset, water exhibits consistently negative relative improvement (-10\% to 0\%) across all shot counts, suggesting fundamental incompatibility with the underlying support tasks. 

\subsubsection{Support Tasks Effects}
\begin{figure*}[h!]
    \centering
    \includegraphics[width=0.95\textwidth]{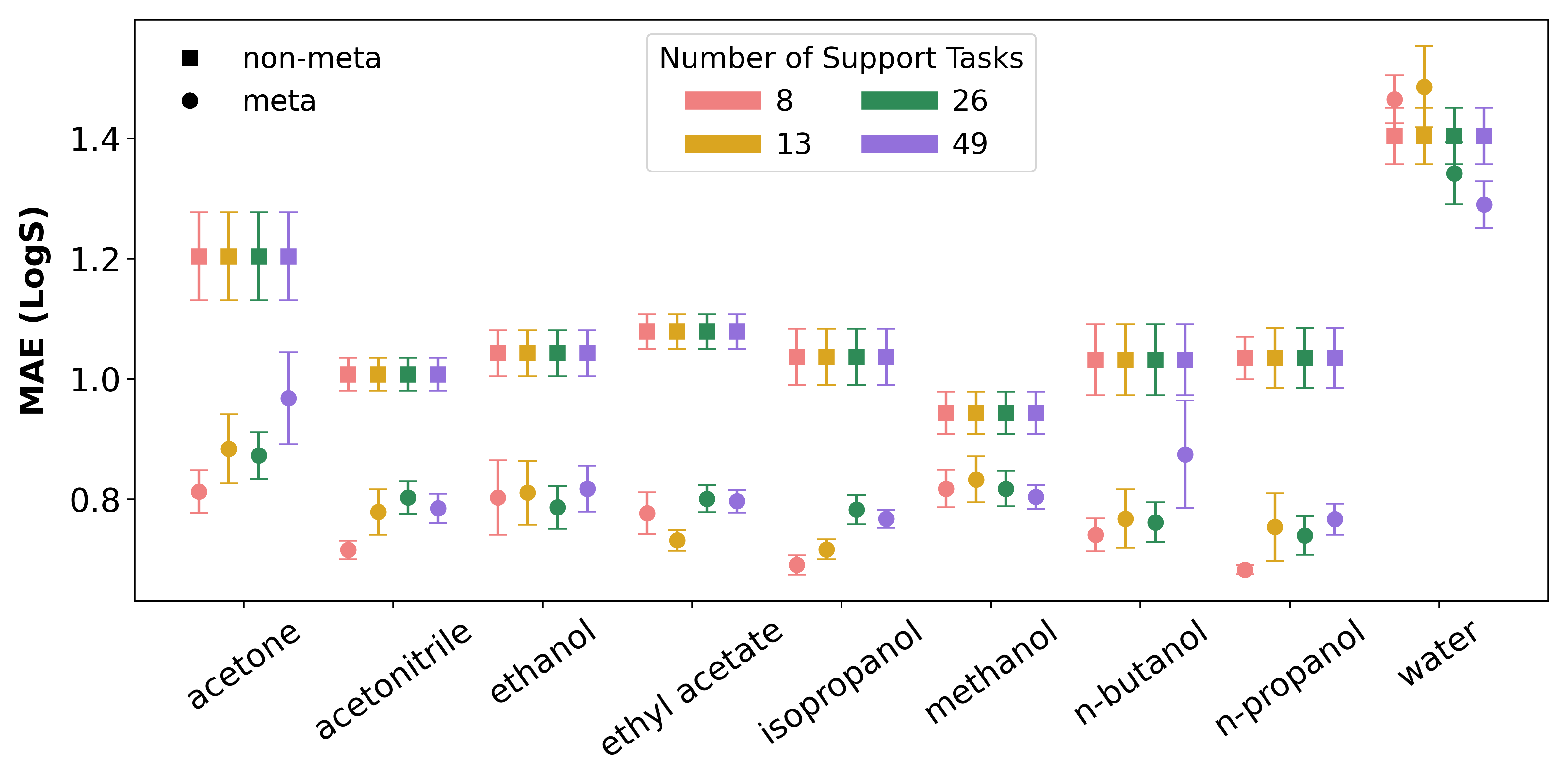}
    \caption{
    Comparison of mean absolute error in predicted solubility across different solvents for meta-learning and vanilla regression models. Results are shown for various numbers of support tasks. Error bars represent standard deviation across experimental runs. Meta-learning consistently achieves lower MAE than non-meta-learning, with performance generally improving as the number of support tasks increases. Maximum support size is set to 5 and 15 shots of the target task was used for specialization.}
    \label{fgr:bsdb_comp_support}
\end{figure*}

To evaluate the impact of support task composition on meta-learning performance, we established four experimental scenarios by varying the minimum datapoints per task (20, 100, 200, and 500), generating task sets containing 50, 27, 14, and 9 tasks respectively. Each configuration employed leave-one-out meta-learning experiments. Fig.~\ref{fgr:bsdb_comp_support} displays MAE values for nine consistent solvents across all task sets, measured at 15 target task training points. The meta-learning approach demonstrates robust superiority over non-meta methods across all solvents except water, with consistently lower MAE values in all four task configurations. We observe that meta-learning variance fluctuates with support task composition, generally showing the lowest MAE in the most data-rich task set ($\geq$500 datapoints/task). This performance enhancement stems from improved support model quality: larger per-task datasets yield more accurate linear regression fits, which subsequently elevate target task prediction accuracy. Our hypothesis is supported by MAE and $R^2$ distributions of individual support models (Fig. S6), where distribution means remain stable across task sets while data-scarce configurations exhibit increased outlier frequency.

\subsection{Solvent Similarity Analysis}

\begin{figure}[h!]
    \centering
    \includegraphics[width=0.85\columnwidth]{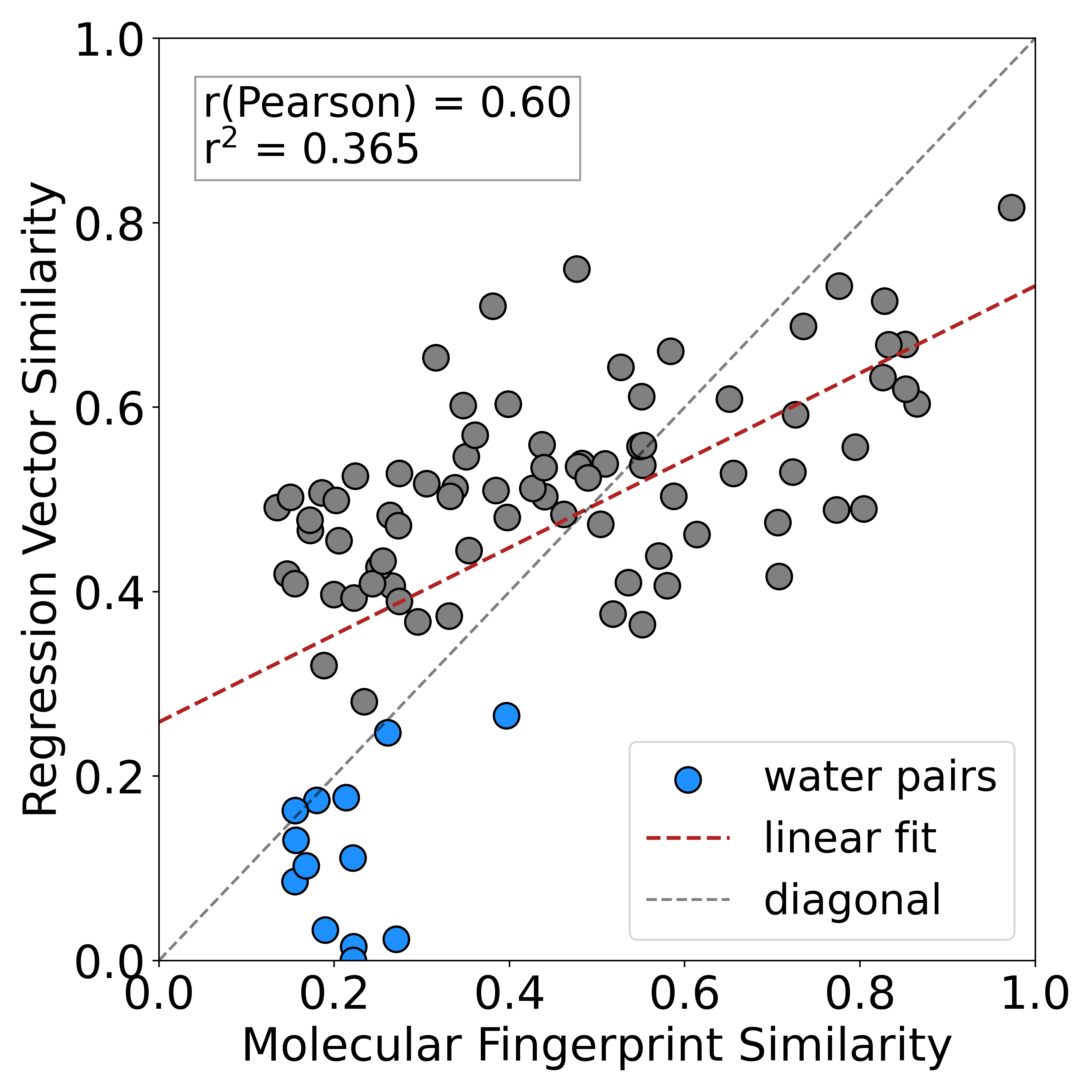}
    \caption{Correlation between molecular fingerprint similarity and regression vector similarity for solvent pairs in BigSolDB 2.0 dataset.
    Each point represents a unique solvent pair. Pearson correlation coefficients and explained variance for linear fits are annotated.}
    \label{fgr:moltask_similarity}
\end{figure}

To better understand the role of task similarity in meta-learning performance for the solubility dataset, we conducted a similarity analysis of the solvent pairs in two different ways. First, we generated topological fingerprints for each solvent molecule using \textit{minervachem}. The cosine similarity measure was calculated between pairs of feature vectors to quantify structural relationships between the solvents.
In parallel, we solved ridge regression models for each of the solvent tasks individually, without meta-learning approaches, taking in all available data for each task and implementing the 80-20 train-test split. From these models, we extracted the regression coefficient vectors, which capture the statistical relationships between molecular fingerprints of the solutes and their solubilities in respective solvent. As earlier, we computed the cosine similarity between each pair of solvent-specific regression vectors. The correlation between two approaches to estimate task similarity is shown in Fig.~\ref{fgr:moltask_similarity} for BigSolDB 2.0. The Boobier et al. dataset demonstrates comparable behavior and can be found in the SI. The moderate Pearson correlations (Boobier et al.:  $R = 0.57$; BigSolDB 2.0: $R = 0.60$) between molecular fingerprint similarity and regression vector similarity reveal consistent alignment between structural features and task-specific solubility relationships across datasets. This agreement suggests both similarity metrics capture complementary aspects of task relationships, with molecular fingerprints providing coarse structural relatedness and regression vectors encoding finer task-specific patterns. Note, in Fig.~\ref{fgr:moltask_similarity} all water-containing pairs group in the lower left corner. Any solvent compared with water shows remarkably low similarity, both if the solvent fingerprints and the individual regression models are compared. The dissimilarity contributes to explaining the lowest meta-learning improvements in aqueous solubility predictions. For meta-learning applications, these trends support the use of similarities to guide initial task grouping to make the most of structural analogies. The correlation consistency across multiple datasets further validates the use of similarity metrics for developing generalizable meta-learning strategies in solubility prediction, particularly in scenarios requiring knowledge transfer between structurally related yet functionally distinct tasks.

The stark contrast between the two similarity matrices—particularly water's extremely low similarity with other solvents in regression space (average of 0.07 with all support tasks)—provides quantitative support for our hypothesis regarding the role of task similarity for the effectiveness of meta-learning.

However, for the Boobier et al. dataset, it is important to consider dataset size alongside chemical similarity of solvents. The total number of data points per solvent varies (Table~\ref{tab:p_sizes}), with water being the solvent with the largest number of observations. Using water as the target reduces the total points available in its support set if compared to other targets. For example, when ethanol is the target, its support tasks contribute 2,348 data points, whereas for water the corresponding total is 1,611. While 1,611 is substantial in many low-data settings, the smaller support set for water—together with water’s chemical dissimilarity to the other solvents—likely contributes to its substantially weaker meta-learning performance. Meta-learning relies primarily on task similarity between the target and support tasks; however, when support tasks are fewer transferable common knowledge might be harder to exploit.

\begin{table}[h]
    \centering
    \renewcommand{\arraystretch}{1.2} 
    \begin{tabular}{cc} 
        \toprule
         & \textit{N datapoints} \\
        \midrule
        water & 1432 \\
        ethanol & 695 \\
        benzene & 464 \\
        acetone & 452 \\
        \bottomrule
    \end{tabular}
    \caption{The sizes of solvent-specific datasets in Boobier et al.}
    \label{tab:p_sizes}
\end{table}

This dual explanation—chemical distinctiveness and limited support dataset size—provides a more nuanced understanding of why water solubility predictions fail to improve with meta-learning. It highlights two critical caveats in applying meta-learning to small-task datasets: 
\begin{enumerate}
    \item task similarity remains a prerequisite for effective knowledge transfer,
    \item small support dataset sizes can negatively affect meta-learning performance.
\end{enumerate}

\subsection{QM9-MultiXC}

\begin{figure}[h!]
    \centering
    \includegraphics[width=0.95\columnwidth]{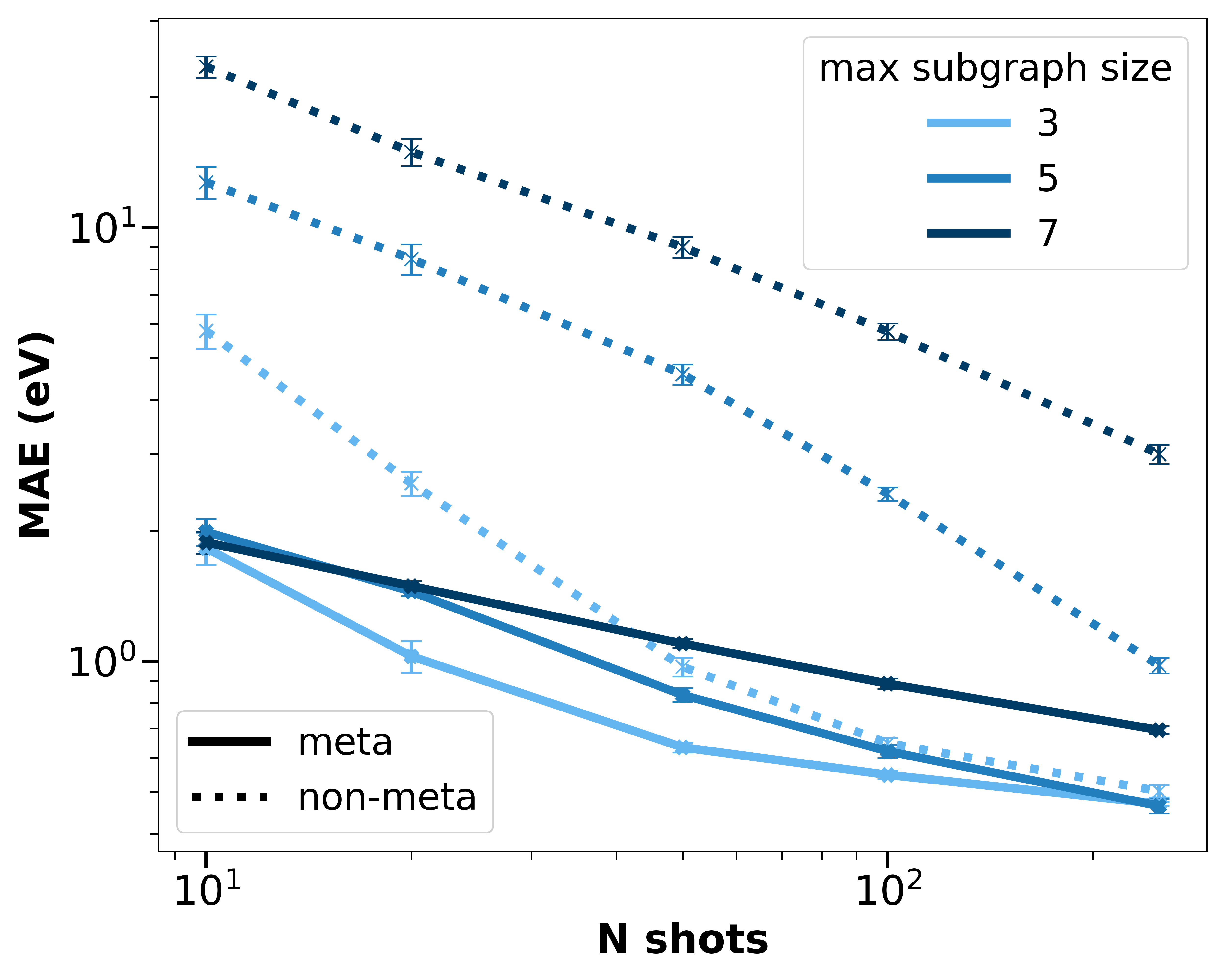}
    \caption{\textit{LAMeL} meta-learning results for predicting molecular atomization enthalpies at MPBE0KCIS\_TZP level of theory using only 5 SZ tasks as support. Performance is presented across three different maximum subgraph sizes. Results are averaged over 10 random initializations with error bars based on the standard deviation of the mean, and all performance metrics are reported for test sets.}
    \label{fgr:tzp_on_sz}
\end{figure}

In contrast to the experimental solubility datasets, the QM9-MultiXC data contains a large number of tasks and an abundance of datapoints per task. The computational methods here can be categorized into three groups based on the basis set used: SZ, DZ, and TZ. It is widely accepted that higher zeta levels provide more accurate results but require more computational resources.~\cite{nagyBasisSetsQuantum2017} As the size of the investigated systems grows, high-zeta basis set calculations become increasingly inaccessible. 
To test the applicability of linear meta-learning to the highly localized chemical properties, we apply our \emph{LAMeL} in various scenarios. First, we investigate whether a limited number of SZ functionals can provide sufficient support for \emph{LAMeL} to predict SZ, DZ, and TZ targets. Additionally, we vary the depth of the fingerprinting process during the featurization step. Unlike the moderate accuracy improvements observed with solubility datasets,  this setup reveals a dramatic reduction in error when applying meta-learning, as evident in Fig.~\ref{fgr:tzp_on_sz}. The most considerable improvements in accuracy were observed in extremely low-shot regimes, where as few as 10 datapoints were used for training on the target task. 
The color gradient in Fig.~\ref{fgr:tzp_on_sz} highlights increasing maximum substructure size, here errors increase and $R^2$ decreases for non-meta approaches. This behavior contrasts with our previous results for solving linear models with \emph{minervachem} topological fingerprinting.~\cite{Tynes2024} However, this is not a true contradiction and arises in this work due to the few-shot nature of experiments.

As maximum substructure size increases, the feature vector size grows significantly (see Table~\ref{tab:parameter_sizes}), while the total number of datapoints remains unchanged. In few-shot regimes, deeper fingerprinting leads to overparameterization given the relatively small molecules in QM9 datasets. For small molecules (no more than nine heavy atoms) a deep featurization can introduce many repeating units that increase non-linearly with depth. It increases dimensionality without adding additional information and worsens ill-conditioning. In high-dimensional settings where the number of features far exceeds the number of observations, even regularized methods like ridge regression can struggle to generalize effectively. For example, training models with 125, 3280, and 82942 parameters on a dataset containing only 50 datapoints leads to significant increases in test error as feature vector complexity grows. In overparameterized scenarios, the model has capacity to capture noise in the training data rather than meaningful patterns only, leading to overfitting. While ridge regression mitigates overfitting by shrinking coefficients, it does not eliminate the problem entirely. This reflects the bias–variance trade-off: with more parameters bias decreases but variance increases substantially, especially in high-dimensional spaces where small perturbations in the data can lead to large changes in predictions. Empirical studies have shown that standard regularization techniques may become less effective in these scenarios unless paired with additional strategies such as dimensionality reduction or adaptive regularization.~\cite{ESL2009, HDdatabook2011}

The results presented in Fig.~\ref{fgr:tzp_on_sz} highlight the interplay between feature complexity and predictive performance in few-shot meta-learning scenarios.  Contrary to the non-meta approach results, the meta-learning framework maintains relatively stable error levels. For the illustrated TZ target, where high-fidelity predictions are inherently more challenging given their no-show among the support tasks, meta-learning achieves comparable MAEs between different substructure depths and demonstrates some resilience against overparameterization by avoiding the sharp fall in performance observed in non-meta models.
The comparative stability of error across substructure sizes in meta-learning highlights its effectiveness in balancing the bias-variance trade-off, even in few-shot regimes with high-dimensional representations. 

\subsubsection{Limited Support Data Investigation}

\begin{figure}[h!]
    \centering
    \includegraphics[width=\columnwidth]{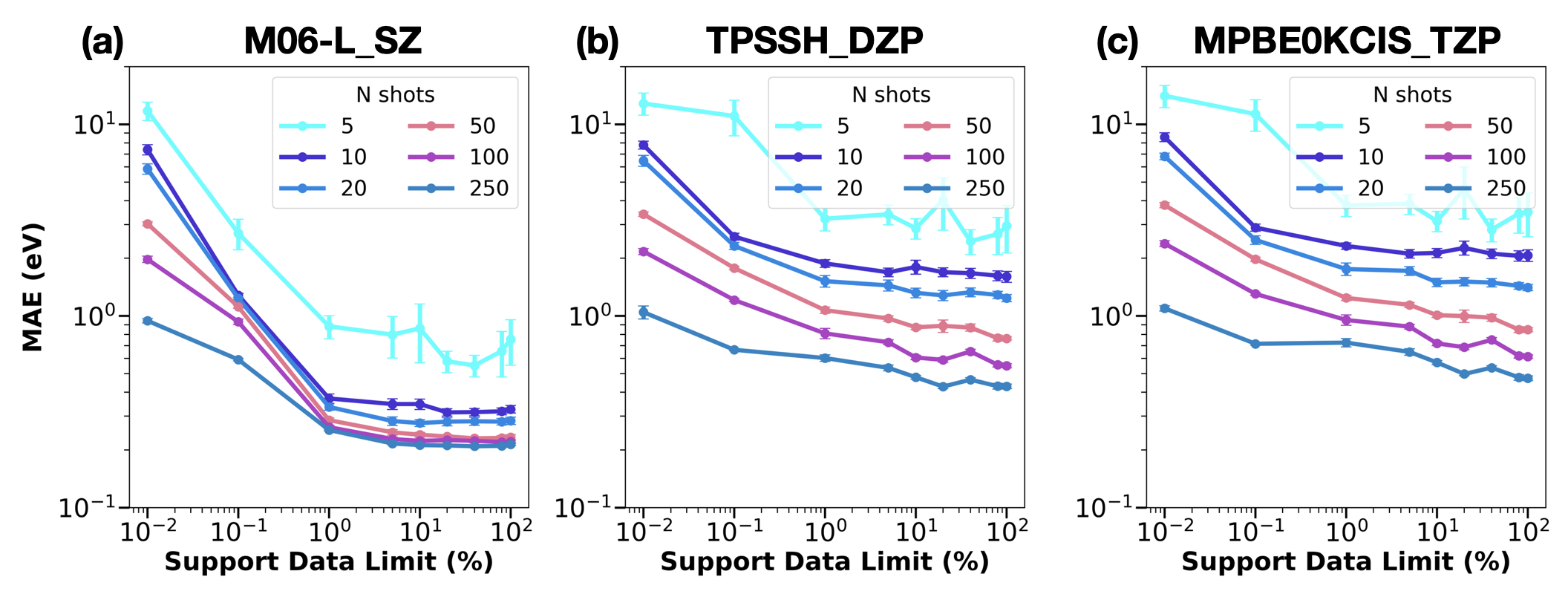}
    \caption{Performance of \textit{LAMeL} in the limited support data regime for three target tasks: (a) M06-L\_SZ, (b) TPSSH\_DZP, and (c) MPBE0KCIS\_TZP. The support data size is represented as a percentage of the total dataset size. The legend indicates the number of shots (NS) used during adaptation to the target task. Only results obtained using \textit{LAMeL} are presented. Maximum substructure size is set to 5.}
    \label{fgr:limdata_meta_szsupport}
\end{figure}

In our experiments with the QM9-MultiXC dataset, we explored the impact of varying support data sizes on meta-learning performance. By default, the support tasks utilized all available datapoints (133,055 per task), providing a comprehensive basis for predictions. To further investigate the effect of support data size, we created subsets at 10, 106, 1064, 5322, 10644, 21288, 42577, 85115 datapoints sampled from the total available support data. 
Interestingly, across all three target tasks (M06-L\_SZ, TPSSH\_DZP, and MPBE0KCIS\_TZP), the relationship between meta-assisted accuracy improvement and support data size remained consistent. As illustrated in Fig.\ref{fgr:limdata_meta_szsupport}, the meta-learning error metrics remained relatively stable for support dataset sizes ranging from 1064 to 106444 datapoints.
This observation suggests that even a small fraction of the large QM9-MultiXC dataset (approximately 1\%) is sufficient to maintain meta-learning efficiency improvements. Nevertheless, when the support sample size dropped below 1\%, error metrics consistently increased across all tested shot sizes for the target task. While meta-learning is robust to reductions in support data size within reasonable limits, extremely small datasets compromise its ability for knowledge extraction. Notably, these findings are different from our observations from solubility datasets, where task similarity played a significant role in determining meta-learning performance. Here, the abundance of data in QM9-MultiXC mitigates some of the challenges posed by task dissimilarity, enabling effective knowledge transfer even with limited support task similarity.

The meta-assisted accuracy improvement depends more on the number of shots of the target task than on the size of the support data. An interesting case arises with the M06-L\_SZ target task, where accuracy improvements from meta-learning show minimal sensitivity to shot size variations (apart from $NS = 5$, which performs noticeably worse). This behavior aligns with our hypothesis regarding task similarity: since all results for this target task were generated using five random SZ-based functionals as support tasks, their inherent similarity facilitates efficient knowledge transfer regardless of shot size.

These results highlight meta-learning's potential for enabling high-fidelity molecular energy predictions using lower-fidelity tasks as support—even under constrained data scenarios—while also emphasizing critical limitations when datasets become extremely sparse.

\section{Conclusions}

In this study, we developed and evaluated a linear meta-learning algorithm for molecular property prediction. Our approach leverages dispersed small-scale data through task-adaptive knowledge transfer without sacrificing interpretability. We assessed performance across three dataset regimes aligned with our objectives: small multitask datasets, large multitask datasets with small per-task data, and large datasets with many tasks and moderate per-task data. We have shown the independent linear meta-learners to use sparse data effectively and improve predictive performance when task similarity is high; under heavy per-task sparsity, meta-learning gains remain, but diminish as similarity between tasks decreases. When both task count and per-task data are high, the proposed method is competitive while retaining interpretability. Overall, our results suggest that dataset structure primarily affects the magnitude of meta-learning benefits, rather than the specific molecular properties (i.e. solubility, atomization energy).

For the solubility datasets, meta-learning yielded up to a 60\% increase in accuracy compared to conventional ridge regression. The magnitude of improvement was closely tied to the degree of similarity among support tasks. In both solubility datasets, water—being the most chemically distinct solvent—stood out as the sole case where meta-learning did not overcome baseline accuracy, highlighting the critical role of support task similarity for successful knowledge transfer. Our results demonstrate that the linear meta-learning framework achieves solubility prediction errors that are on par with those reported for deep learning models. For the nine most popular solvents in the BigSolDB 2.0 dataset (Fig.~\ref{fgr:bsdb_comp_support}), the mean absolute error (MAE) can be consistently reduced to below 0.800 LogS units, with the lowest MAE observed at 0.683$\pm$0.007 for n-propanol. This level of accuracy is comparable with literature: Ulrich et al. report an experimental uncertainty of 0.5–0.6 log units and an ML model with RMSE of 0.657 for aqueous solubility,~\cite{ulrichPredictionWaterSolubility2025} MolMerger achieves an average MAE of 0.79 LogS units across solute–solvent pairs,~\cite{MolMerger} AttentiveFP~\cite{AttentiveFP} and MoGAT~\cite{MoGAT023}, both limited to aqueous systems, report RMSE values of 0.61 and 0.478 log units, respectively, while SolPredictor~\cite{ahmadSolPredictorPredictingSolubility2024} reaches an average RMSE of 1.09 log units for aqueous solubility. The ability of our linear meta-learning approach to deliver comparable predictive performance across a chemically diverse set of solvents supports its practical utility in real-world solubility prediction tasks with minimal available data.

For the atomization energy dataset, which involves highly localized electronic properties, linear meta-learning provided the largest relative gains, further supporting the applicability of the method to various tasks and settings. Our study demonstrates the data efficiency achieved by the meta-learning framework: accurate predictions were obtained using as little as 1\% (i.e. 1064 datapoints per support task) of the full training data in the QM9-MultiXC dataset, demonstrating the potential of this method for scenarios where data collection is expensive or time-consuming. These findings suggest that meta-learning not only interpolates between tasks but also captures underlying physical and chemical principles, enabling interpretative extrapolation even in low-data regimes.

While the linear nature of the model constrains its capacity to capture complex relationships, its simplicity allows for robust and interpretable performance. Future work should explore the integration of nonlinear meta-learners for interpretability, the extension of linear meta-learning approach to more chemically diverse and challenging systems, and the incorporation of active learning strategies to further enhance data efficiency and predictive power.

Overall, our results establish linear meta-learning as a powerful and computationally efficient paradigm for molecular property prediction. By enabling significant accuracy gains with minimal data, this approach holds promise for accelerating high-throughput screening and materials discovery, particularly in domains where experimental resources are limited and quick adaptation is essential.

\begin{acknowledgement}
We acknowledge the support by the U.S. Department of Energy, Office of Science, Office of Basic Energy Sciences, Heavy Element Chemistry Program at Los Alamos National Laboratory (LANL) (Y.P., P.Y., N.L.) (contract no KC0302031 LANL2023E3M2). We gratefully acknowledge the support of the U.S. Department of Energy through the LANL Laboratory Directed Research Development Program under project number 20250637DI for this work (Y.P., N.L.). This research used resources provided by the CAI-1 Darwin HPC cluster at LANL. Los Alamos National Laboratory is operated by Triad National Security, LLC, for the National Nuclear Security Administration of the U.S. Department of Energy (contract no. 89233218CNA000001).  
\end{acknowledgement}

\bibliography{mlearner_refs}

\onecolumn
\appendix
\section{Supporting Information}
\section{Boobier et al. all results}
\subsection{Individual solvents solubilities under varying maximum substructure size}

\begin{figure}[h!]
    \centering
    \includegraphics[width=0.95\columnwidth]{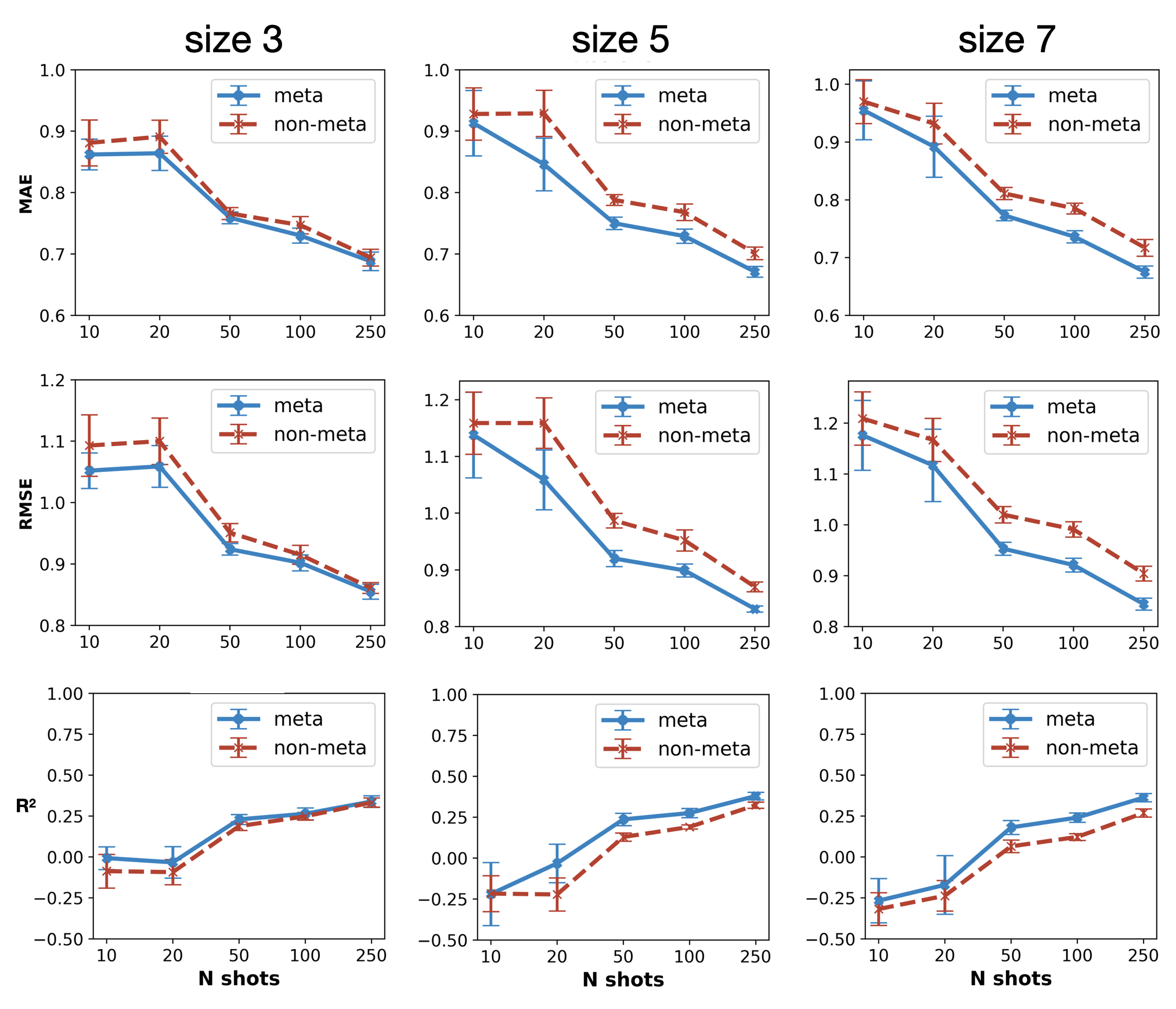}
    \caption{\textit{LAMeL} meta-learning results for predicting solubilities of small organic molecules in acetone with maximum substructure size 3, 5, 7 (columns) being used in the fingerprinting process. The data is averaged over 10 random initializations.}
    \label{fgr:boobier_acetone}
\end{figure}

\begin{figure}[h!]
    \centering
    \includegraphics[width=0.95\columnwidth]{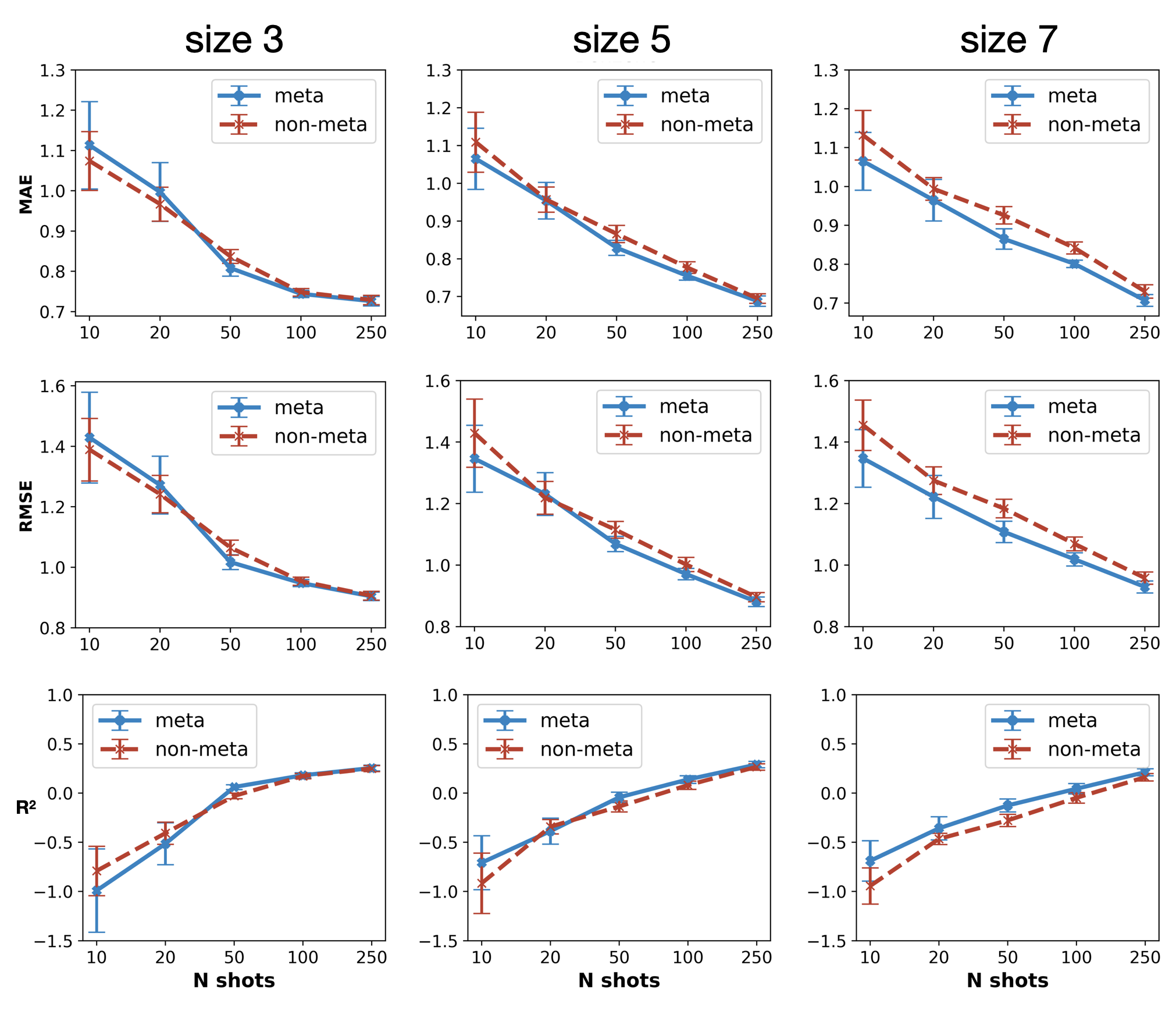}
    \caption{\textit{LAMeL} meta-learning results for predicting solubilities of small organic molecules in benzene with maximum substructure size 3, 5, 7 (columns) being used in the fingerprinting process. The data is averaged over 10 random initializations.}
    \label{fgr:boobier_benzene}
\end{figure}

\begin{figure}[h!]
    \centering
    \includegraphics[width=0.95\columnwidth]{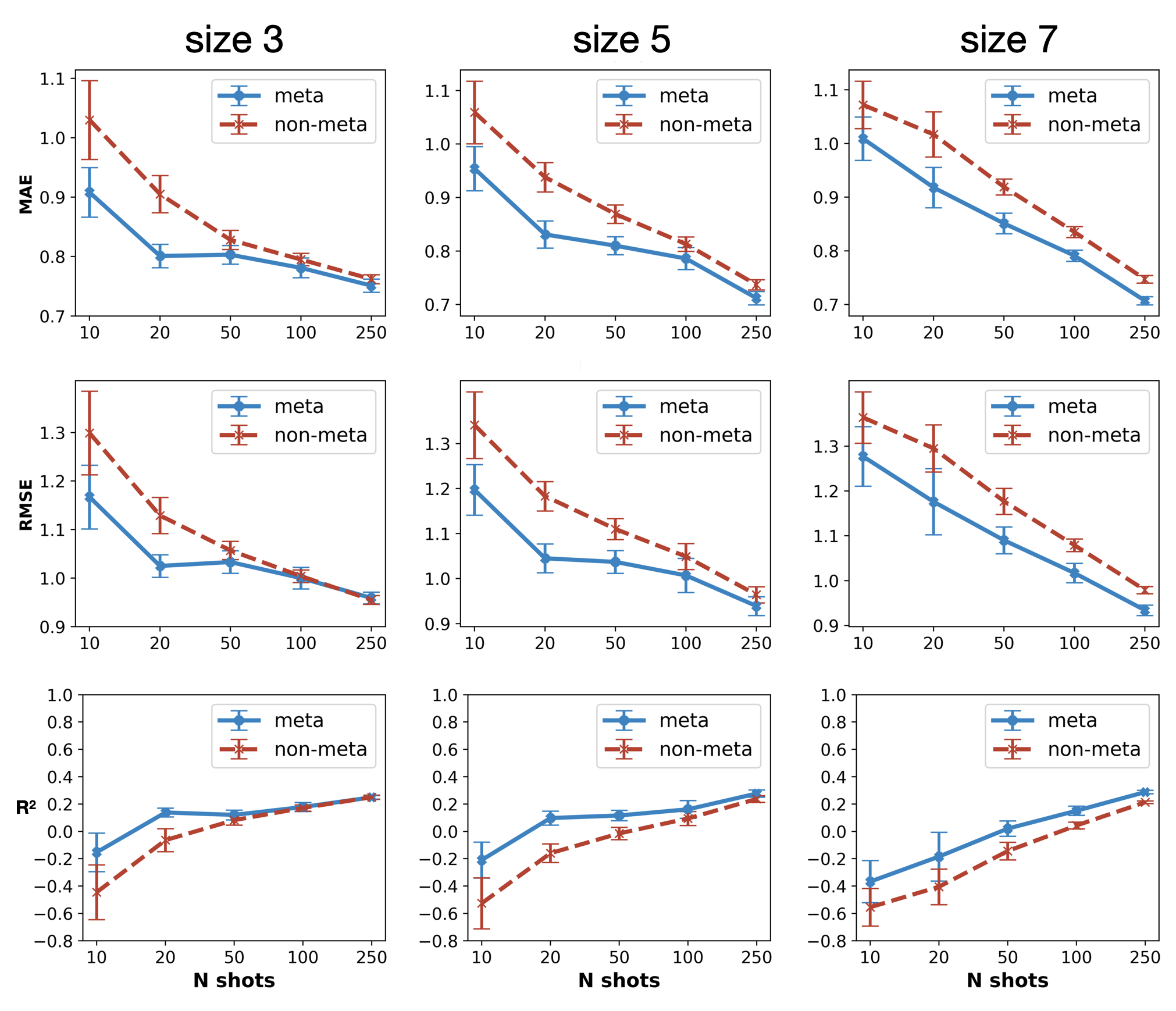}
    \caption{\textit{LAMeL} meta-learning results for predicting solubilities of small organic molecules in ethanol with maximum substructure size 3, 5, 7 (columns) being used in the fingerprinting process. The data is averaged over 10 random initializations.}
    \label{fgr:boobier_ethanol}
\end{figure}

\begin{figure}[h!]
    \centering
    \includegraphics[width=0.95\columnwidth]{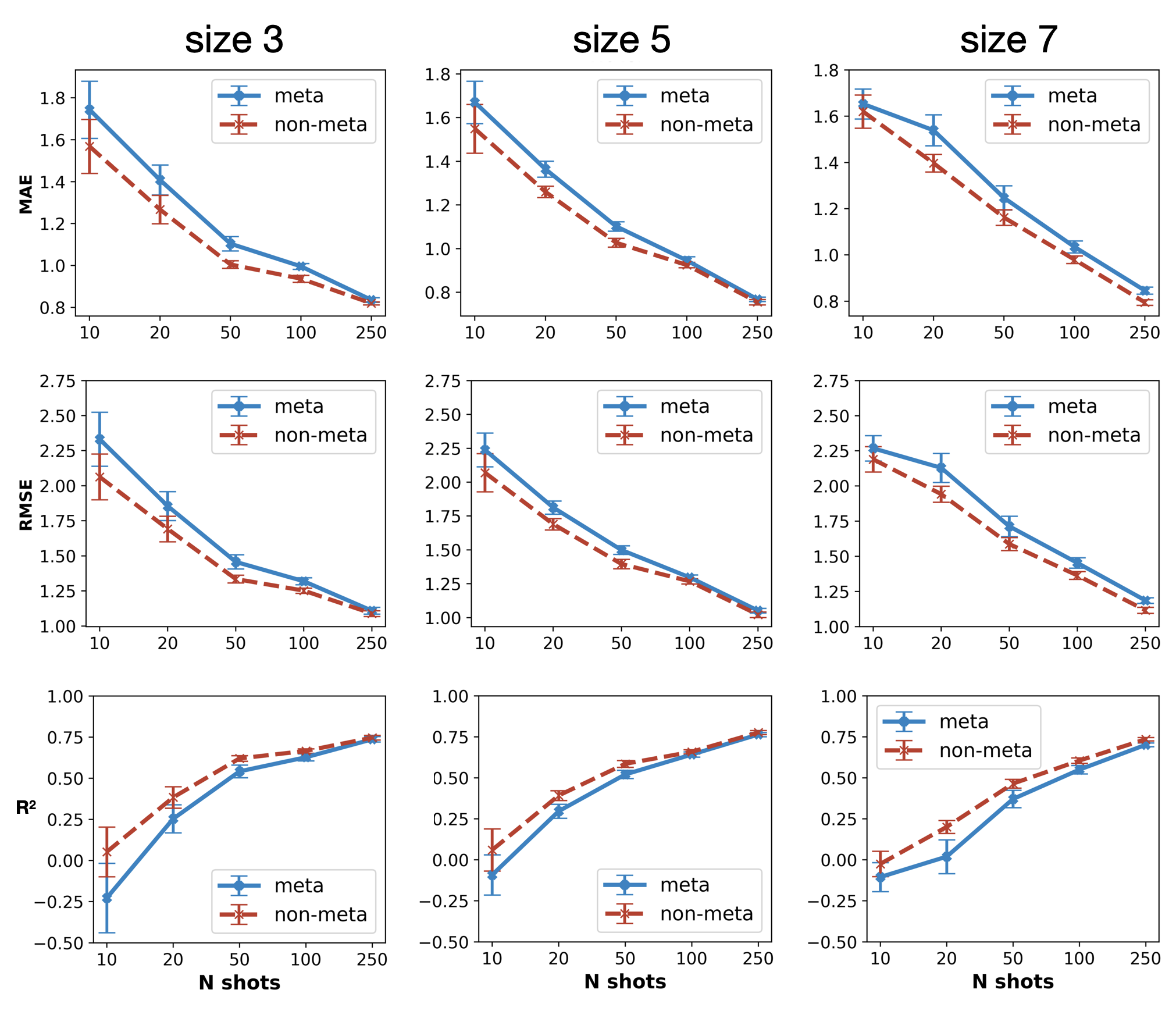}
    \caption{\textit{LAMeL} meta-learning results for predicting solubilities of small organic molecules in water with maximum substructure size 3, 5, 7 (columns) being used in the fingerprinting process. The data is averaged over 10 random initializations.}
    \label{fgr:boobier_water}
\end{figure}
\clearpage
\section{BigSolDB2.0 results}

\begin{figure}[h!]
    \centering
    \includegraphics[width=0.5\columnwidth]{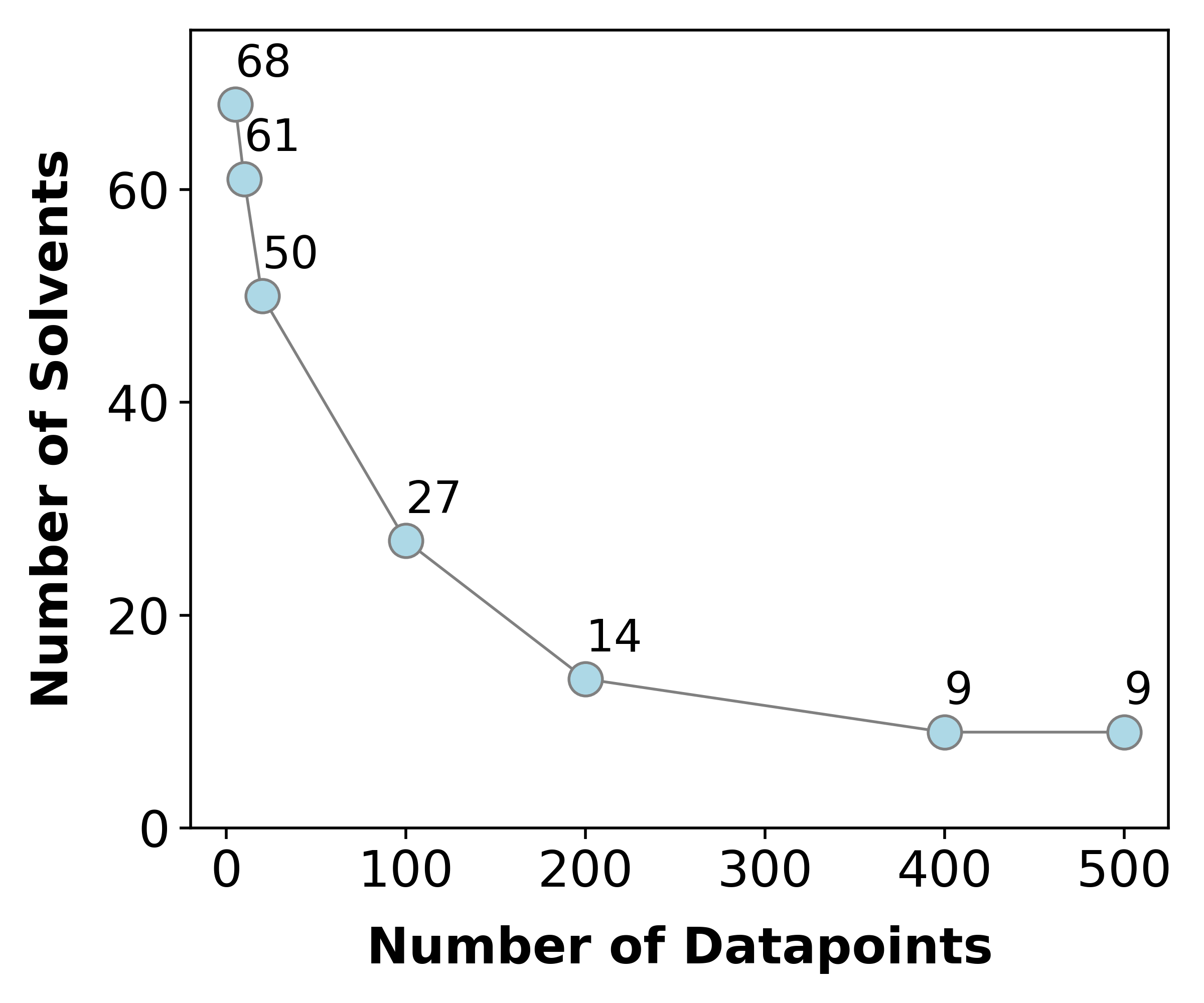}
    \caption{Relationship between the limits on the number of datapoints and the number of available solvents.}
    \label{fgr:solvents_vs_datasize}
\end{figure}

The relationship in Fig.~\ref{fgr:solvents_vs_datasize} shows that as the minimum required number of datapoints per solvent increases, the number of solvents available for analysis decreases sharply.

\begin{figure}[h!]
    \centering
    \includegraphics[width=0.7\columnwidth]{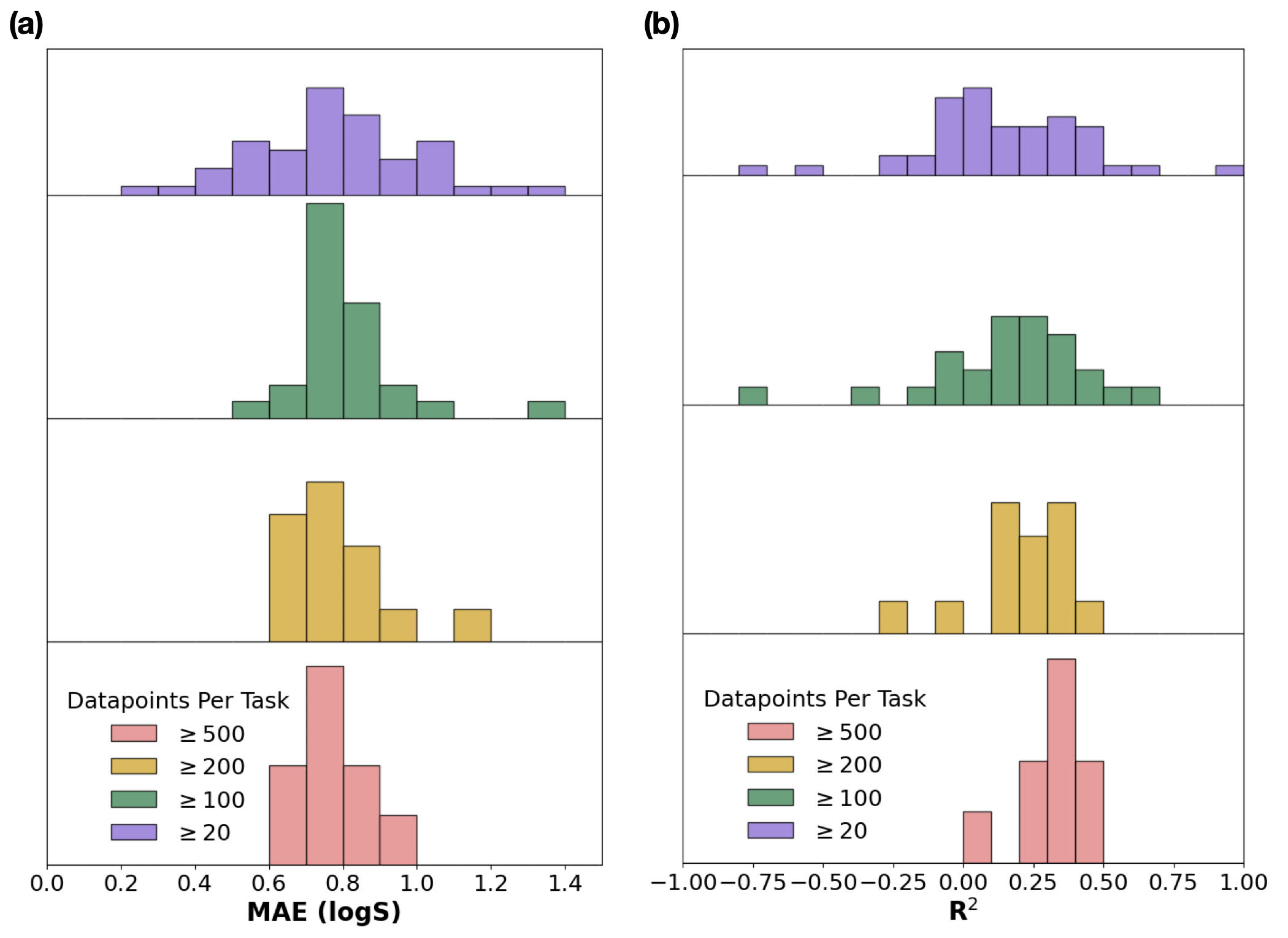}
    \caption{The distribution of performance metrics MAEs (a) and R$^2$ (b) for the individual independent ridge regression models depending on the size limitation imposed on solvent-specific datasets.}
    \label{fgr:base_models}
\end{figure}
\clearpage
\section{QM9-MultiXC results}

\begin{figure}[h!]
    \centering
    \includegraphics[width=\columnwidth]{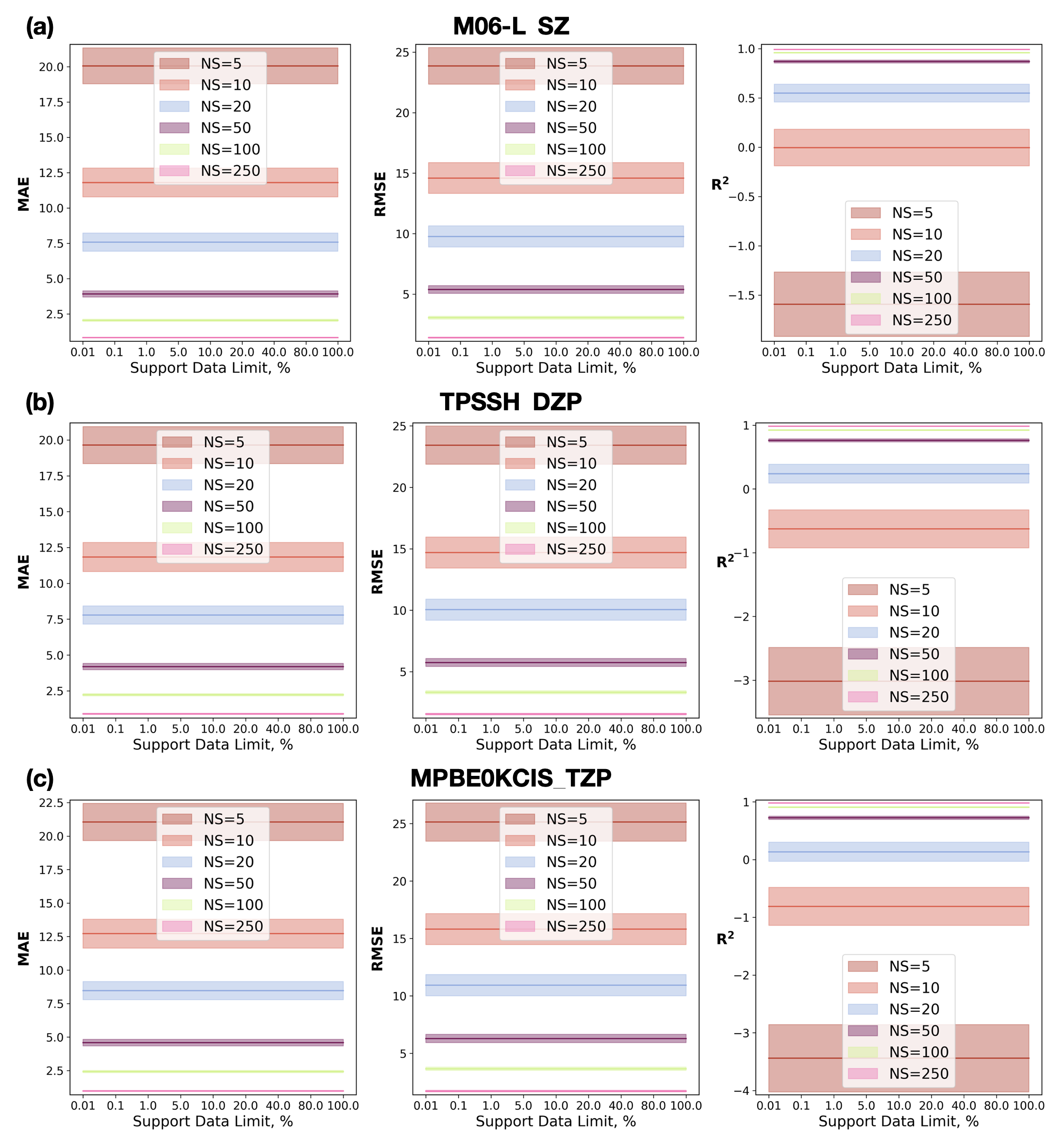}
    \caption{Non-meta ridge regression performance for three target tasks: (a) M06-L\_SZ, (b) TPSSH\_DZP, and (c) MPBE0KCIS\_TZP. The legend indicates the number of shots (NS) used during training. }
    \label{fgr:limdata_nonmeta_szsupport}
\end{figure}

\end{document}